\documentclass[10pt,journal,compsoc]{IEEEtran}

\ifCLASSOPTIONcompsoc
  \usepackage[nocompress]{cite}
\else
  \usepackage{cite}
\fi

\ifCLASSINFOpdf
\else
\fi

\usepackage[dvipsnames]{xcolor}
\usepackage{xr}

\usepackage{microtype}     %
\PassOptionsToPackage{warn}{textcomp}  %
\usepackage{textcomp}      %
\usepackage{mathptmx}      %
\usepackage{times} %
\usepackage{cite}  %
\usepackage{tabu}  %
\usepackage{booktabs}      %

\usepackage{subfigure}
\usepackage[export]{adjustbox}
\usepackage{hyperref}

\usepackage[inline]{enumitem}
\usepackage{amssymb}
\usepackage{amsmath}
\usepackage{bm}
\usepackage[normalem]{ulem}

\newcommand{\etal}{\textit{et\penalty50\ al}.}
\newcommand{\ie}{\textit{i}.\textit{e}.,~}
\newcommand{\eg}{\textit{e}.\textit{g}.,~}

\newlist{inlinelist}{enumerate*}{1}
\setlist*[inlinelist,1]{%
  label=(\arabic*),
}

\newcommand{\revA}[1]{#1}

\newcommand{\revB}[1]{#1}
\newcommand{\revBa}[1]{#1}

\newcommand{\revC}[1]{#1}

\newcommand{\revD}[1]{#1}

\newcommand{\revDg}[1]{#1}

\newcommand{\revDp}[1]{#1}

\newcommand{\revDr}[1]{}

\newcounter{row}
\newcommand{\nextrow}{\refstepcounter{row}\arabic{row}}

\newcolumntype{R}{>{\global\let\currentrowstyle\relax}}
\newcolumntype{B}{>{\currentrowstyle}}
\newcommand{\rowstyle}[1]{\gdef\currentrowstyle{#1}%
    #1\ignorespaces
}

\usepackage{scrlayer-scrpage}
\clearpairofpagestyles %

\DeclareNewLayer[
  foreground, %
  headsep, %
  voffset=0.1cm, %
  contents={%
    \makebox[\layerwidth]{%
      \parbox{\dimexpr\layerwidth-1cm\relax}{%
        \centering%
        \scriptsize %
        This article has been accepted for publication in IEEE Transactions on Visualization and Computer Graphics. This is the author's version which has not been fully edited and content may change prior to final publication. Citation information: DOI \href{https://doi.org/10.1109/tvcg.2023.3313729}{10.1109/TVCG.2023.3313729}%
      }%
    }%
  }
]{headerbox}

\DeclareNewLayer[
  foreground, 
  align=b, 
  voffset=\paperheight, %
  height=3cm, %
  contents={%
    \makebox[\layerwidth]{%
      \parbox{\dimexpr\layerwidth-1cm\relax}{%
        \centering%
        \scriptsize
        © 2023 IEEE. Personal use is permitted, but republication/redistribution requires IEEE permission. See https://www.ieee.org/publications/rights/index.html for more information.
      }%
    }%
  }
]{footerbox}

\AddLayersToPageStyle{@everystyle@}{headerbox, footerbox}

\begin{document}

\title{Reinforced Labels: Multi-Agent \\Deep Reinforcement Learning \\for Point-\revC{F}eature Label Placement}
\author{
    Petr Bob\'{a}k, 
    Ladislav \v{C}mol\'{\i}k, 
    Martin \v{C}ad\'{\i}k
    \IEEEcompsocitemizethanks{
        \IEEEcompsocthanksitem P. Bob\'{a}k and M. \v{C}ad\'{\i}k are with \revDp{the} Faculty of Information Technology, Brno University of Technology. E-mail: \{ibobak\,$|$\,cadik\}@fit.vutbr.cz.
        \IEEEcompsocthanksitem L. \v{C}mol\'{\i}k 
        is with \revDp{the} Faculty of Electrical Engineering, Czech Technical University in Prague. E-mail: 
        cmolikl%
@fel.cvut.cz.
    }
}

\IEEEtitleabstractindextext{%
\begin{abstract}
\revC{Over \revDp{the} recent years}, Reinforcement Learning combined with Deep Learning techniques has successfully proven to solve complex problems in various domains, including robotics, self-driving cars, and finance. 
In this paper, we are introducing Reinforcement Learning (RL) to \revC{\textit{label placement}, a complex task in data visualization that seeks optimal positioning for labels to  avoid overlap and ensure legibility}. 
Our novel point-feature label placement method utilizes Multi-Agent Deep Reinforcement Learning to \textit{learn} the label placement strategy, the first machine-learning-driven labeling method, in contrast to \revDp{the} existing hand-crafted algorithms designed by human experts.
To facilitate RL learning, we developed an environment where an agent acts as a proxy for a label, a short textual annotation that augments visualization. 
Our results \revC{show} that the strategy trained by our method significantly outperforms the random strategy of an untrained agent \revC{and \revDp{the} compared methods designed by human experts} in terms of \textit{completeness} \revB{(\ie the number of placed labels). 
The trade-off is increased computation time, making the proposed method slower than \revDp{the} compared methods. %
Nevertheless, our method is ideal for \revC{scenarios} where the labeling can be computed in advance, and completeness is essential, such as cartographic maps, technical drawings, and medical atlases.}
\revB{Additionally, we conducted a user study to assess the perceived performance. The outcomes revealed that the participants considered the proposed method to be significantly better than the other examined methods. This indicates that the improved completeness is not just reflected in the quantitative metrics but also in the subjective evaluation \revDr{of}\revDp{by} the participants.}
\end{abstract} %

}

\maketitle
\thispagestyle{plain}

\IEEEdisplaynontitleabstractindextext

\IEEEpeerreviewmaketitle

\IEEEraisesectionheading{\section{Introduction}\label{sec:introduction}}

\IEEEPARstart{T}{extual} \textit{labels} on top of a visualization serve as an additional informational layer (\revDr{so-called}\revDp{known as} \textit{labeling}) to features of interest. An example of such a visualization could be a geographical map where cities (\textit{point features}), highways (\textit{line features}), and bodies of water (\textit{area features}) are the features of interest. 
However, according to Yoeli~\cite{Yoeli1972}, a founder of automated label placement, manual labeling of geographical maps takes up to 50\% of the overall production time. On the other hand, even simplified versions of \revDg{the} label placement \revC{problem} are NP-hard~\cite{Marks1991}. Therefore, it is often
necessary to use heuristics in real-world scenarios to find feasible positions for a large number of labels.

The automatic label placement was acknowledged by ACM Computational Geometry Impact Task Force~\cite{Chazelle1999} as an important area of research more than two decades ago. In this work, we focus on automatic point-feature label placement \revA{(PFLP) or, concisely and interchangeably, point-feature labeling (PFL).}
\revA{Over the past decades, }many distinct techniques were published: techniques based on mathematical programming~\cite{Zoraster1986b}, simulated annealing~\cite{Christensen1995}, tabu search~\cite{Yamamoto2002b, Alvim2009}, evolution and genetic algorithms~\cite{Lu2019}, %
force-based algorithms\cite{Hirsch1982}, and rule-based algorithms\cite{Mote:IV:2007, Luboschik2008, Kittivorawong20, Pavlovec2022}.
These techniques share one commonality -- all of them are hand-crafted algorithms designed by human experts. 
Mathematical programming-based approaches often suffer from extensive computation time, while objectives are challenging to optimize due to solver limitations (\eg linearity, integral or quadratic constraints), and optimality is only guaranteed in limited cases. As a result, hand-engineered objectives  may not describe the actual objective function of PFL in the sense of mathematical formulation.
Rule-based techniques are greedy algorithms, that once the solution for a label is found, its position cannot be rearranged later. Therefore, they do not guarantee optimality and\revDr{. These techniques} are less flexible by design since objectives have to be hardcoded in the algorithm.
Evolution, genetic, and force-based approaches often get stuck in local optima.

Just as deep learning vastly mitigated the need for feature engineering and produced remarkable performance improvements in computer vision and natural language processing, applying learning techniques to visualization could lead to similar advancements.

In this work, we present point-feature label placement as a \textit{reinforcement learning} problem. 
Reinforcement Learning (RL) is an area of machine learning (alongside supervised and unsupervised learning) concerned with decision-making, driven by experience, to maximize the numerical reward signal. Over the discrete-time steps, an agent (\ie decision-maker) senses the \textit{state} of the \textit{environment}, interacts with the environment by taking \textit{actions} that affect\revDr{s} the state, and receives a \textit{reward}. 
Learning from experience overcomes the lack of label placement datasets that would be needed for supervised learning.
Furthermore, the advancement in Deep Learning and its combination with RL emerged in\revDr{to} Deep Reinforcement Learning (DRL). DRL has enabled scaling the RL to previously intractable problems (includ\revC{ing} labeling) due to the curse of dimensionality. Recent research has successfully \revDr{proved}\revDp{proven} that DRL can solve complex problems within several domains even with enormous state spaces, \eg robotics~\cite{Levine2016, Levine2017},\revDr{ video games~\cite{Mnih2013, Silver2016},} self-driving cars~\cite{Pan2017}, industrial design~\cite{Mirhoseini2021}\revDp{,} and finance~\cite{Deng2017}. %
Moreover, DRL can be trained for objectives that are difficult to optimize directly, as DRL is agnostic to the precise model of the environment as long as the reward signal correlates with the objective. Li and Malik~\cite{Li2017} demonstrated that an RL-based autonomous optimization algorithm converges faster and/or finds better optima than \revDp{the} existing hand-engineered optimization algorithms (\eg gradient descent, momentum, conjugate gradient).
\revDr{Additionally, the DRL agent can be fine-tuned~\cite{Mirhoseini2021} for a specific task (\eg specific set of points of interest) or imitate the behavior of the domain expert~\cite{Codevilla2018} (\eg cartographer, illustrator).}

Given the mentioned characteristics, we believe the DRL approach is well suited for the point-feature label placement problem. \revD{However, employing RL to label placement poses unique challenges, such as complex deep RL training, handling variable number of anchors and labels, and vast continuous state and action spaces. 
Moreover, it demands the meticulous design of neural network architectures, formulation of state space, and careful definition of action space. Despite these challenges, the potential of RL in this domain is immense. Our research confronts these issues, devising strategies highlighting our novel approach to the problem.}

To the best of our knowledge, no work has been published on label placement with the ability to learn and generalize from experience and generate new labeling for unseen sets of features of interest.
Our main contributions are summarized as follows:

\begin{enumerate}[label=(\arabic*)]
    \item \revB{We introduce a Multi-Agent Deep Reinforcement Learning formulation to the point-feature labeling problem, which we believe to be the first machine-learning-driven labeling method contrary to \revDp{the} existing hand-crafted algorithms designed by human experts. 
    Our efficient feedforward neural network architecture, with less than half a million parameters, serves as both a policy and value function approximation.}
    \item \revC{We provide comprehensive ablation studies on \revDp{the} observed modalities and neural network architecture that justify our design choices, highlight \revDp{the} essential aspects of our work, and provide a guideline for future work. As a part, we introduce a novel \textit{completeness metric} that measures the performance of labeling methods based on the number of completely labeled point features and \textit{benchmark dataset}\revDr{that serves as a substrate} for evaluation.}
    \item \revC{We compare the performance of the proposed method with two existing methods, Particle-Based Labeling~\cite{Luboschik2008} and Rapid Labels~\cite{Pavlovec2022}, using quantitative assessments. Additionally, we conducted a user study to evaluate the proposed method qualitatively. Based on these evaluations, we show that our method outperforms all\revDr{ of} the other examined methods.}
\end{enumerate}

\section{Related Work}
\label{sec:related-work}

\begin{figure}[t]
    \subfigure[] {
    \label{fig:fixed-4-model}
    \includegraphics[width=0.145\textwidth]{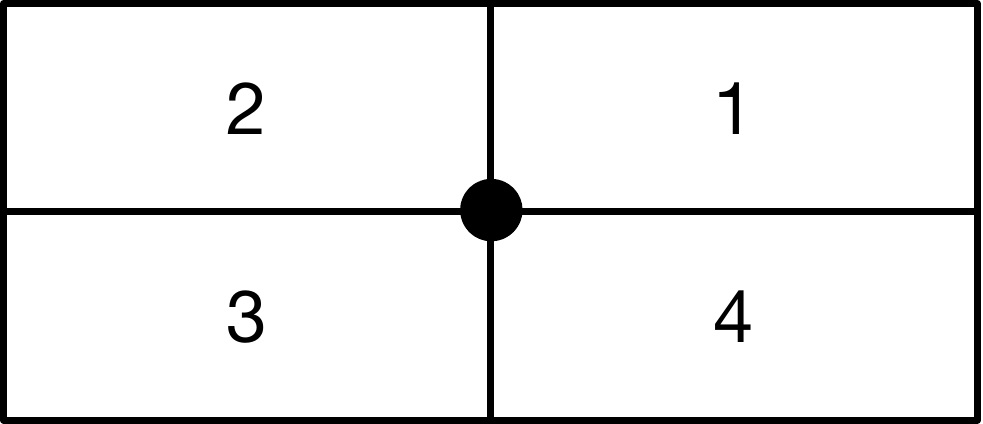}
    }
    \hfill
    \subfigure[] {
    \label{fig:fixed-8-extension-model}
    \includegraphics[width=0.145\textwidth]{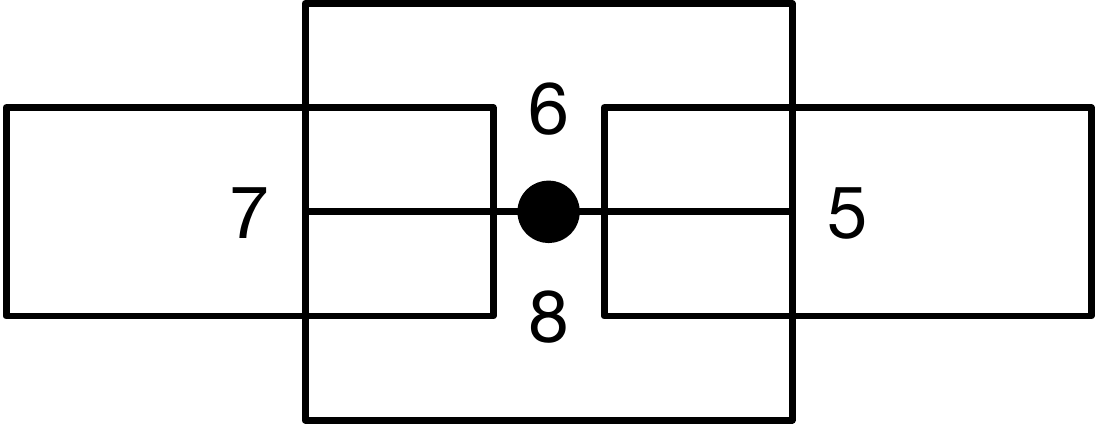}    
    }
    \hfill
    \subfigure[] {
    \label{fig:sliding-4-model}
    \includegraphics[width=0.145\textwidth]{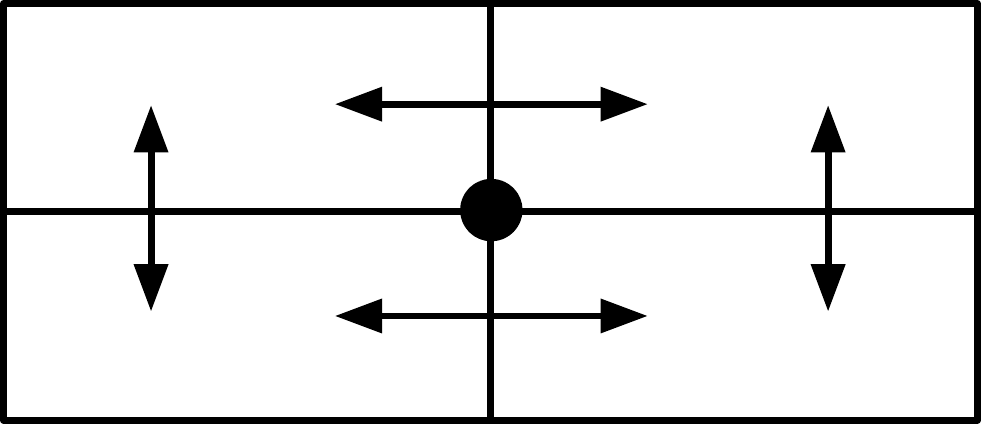}
    }
    \caption{Illustration of label placement models according \revDp{to} \revA{Luboschik~\etal~\cite{Luboschik2008}}. Image \subref{fig:fixed-4-model} depicts \revDg{the} 4-position model, \subref{fig:fixed-8-extension-model} shows the extended positions of \revDg{the} 8-position model, and  \subref{fig:sliding-4-model} displays \revDg{the} slider model. The number of each label position indicates the rank where one is the most preferred position and eight is the least preferred one.}
    \label{fig:internal-models}
\end{figure}

\revC{%
In our review of related work, we adopt the classification structure presented by Luboschik~\etal~\cite{Luboschik2008}. We first discuss point-feature labeling approaches that utilize the prevalent \textit{fixed-position} model. These techniques make use of a few predetermined candidate locations in close proximity to each point feature and select one of these as the label position. Subsequently, we delve into methods that employ the \textit{slider} model, allowing for more flexible label placement by permitting the label to slide around its associated point feature. Illustrations of the fixed-position and slider models can be found in \autoref{fig:internal-models}. The labels in fixed-position and slider models are denoted by Luboschik~\etal~\cite{Luboschik2008} as \textit{adjacent} labels as they are positioned adjacent to the point feature.   
Only a few approaches~\cite{Zoraster1997, Luboschik2008}  have gone a step further to enhance the flexibility of the fixed-position or slider model by introducing \textit{distant labels}. These labels are permitted to be placed further away from the associated point features while the association is maintained via lines or curves.
In the following text, we explicitly mention which bodies of work allow distant labels.
As our work harnesses reinforcement learning to address the point feature labeling problem, we also explore visualization approaches that leverage machine learning.
}

\subsection{Fixed-position Models}
The methods based on fixed-position models are restricted in terms of possible label positions, and thus the search space is considerably narrowed. \revB{The goal is to place labels for as many point features as possible by choosing from the pre-determined candidates. Often, rules are applied to simplify the resolution of conflicts between the candidates~\cite{Wagner2001}.} The optimal label layout (\ie the label layout with the maximum number of placed labels) is often not achieved because the optimal position of labels is not present in the discrete space of the fixed-position model. Nevertheless, the fixed-position models have been studied \revDr{in much greater detail}\revDg{more extensively} than the slider models.

\revB{Various optimization strategies have been employed to find the 
label placement.} 
Zoraster~\cite{Zoraster1986b} proposed an integer linear programming formulation of label placement in maps specific to \revDp{the} oil industry. %
\revC{Christensen and Marks~\cite{Christensen1995} and Zoraster~\cite{Zoraster1997} proposed 
methods based on simulated annealing. The approach of Zoraster expands the fixed-position model to incorporate distant labels. Each distant label is chosen from candidates that are located in 
the same directions from the point feature as in the fixed-position model but at two additional distance levels.}
Yamamoto~\etal~\cite{Yamamoto2002b} introduced the combinatorial optimization method based on tabu search. %
Alvim~\etal~\cite{Alvim2009} proposed a point-feature labeling method based on the application of the POPMUSIC (Partial Optimization Metaheuristic under Special Intensi\revDp{fi}cation Conditions)\revDr{ metaheuristic}. The POPMUSIC\revDr{ metaheuristic} divides the PFL into subproblems that are optimized with tabu search. %
Rabello~\etal~\cite{Rabello2014} presented a clustering search metaheuristic\revDr{The CS can detect promising areas of the search space, and prevent more intense search in poor areas or areas that have already been sufficiently explored.}\revDg{, which identifies promising search areas while avoiding over-exploring poor or well-explored regions.} %
Lu~\etal~\cite{Lu2019} proposed a hybrid approach combining discrete differential evolution and genetic labeling algorithm for all types of features (\ie point, line, area). %
While all the above-described methods provide high-quality label layouts, they share a common problem. 
They require a significant computational effort, which results in high runtimes. 

\revDr{Greedy methods, in contrast to optimization methods,}\revDp{Unlike optimization methods, greedy methods} \revA{reduce the runtime by sequentially labeling the point features without the ability to recover from bad solutions or with a limited recovery step at the end~\cite{yamamoto2005fast}.}
Typically, greedy methods will find a worse label layout but in \revDg{a} significantly shorter time (milliseconds)
which makes them usable in time\revDg{-}critical applications.
Mote~\cite{Mote:IV:2007} introduced a greedy method that divides the screen into a 2-dimensional grid and determines a simplified version of a conflict graph of the point features. Only point features in neighboring cells need to be checked for conflict with a given point feature. %
Luboschik~\etal~\cite{Luboschik2008} proposed a greedy particle-based labeling (PBL) approach that can also respect other visual elements and the visual
extent of labeled features using conflict particles %
representing the occupied space.
Each placed label is approximated by a set of
newly generated particles. The performance of PBL depends heavily on the number of used particles. The method is significantly slower for cases with complex visual features represented by many particles.
Kittivorawong~\etal~\cite{Kittivorawong20} addressed the performance problem of particle-based labeling with
an occupancy bitmask that allows faster evaluation of label overlaps with complex visual features that cannot be occluded. Nevertheless, the computation time required to determine overlap for a label still depends on the size of the label and the resolution of the screen.
Recently, Pavlovec and \v{C}mol\'{\i}k~\cite{Pavlovec2022} introduced Rapid labels, an approach leveraging the power of GPU. They are allowing to label several point features in each iteration using a 2-dimensional grid. Further, they evaluate the overlaps of labels with important visual elements, the conflicts between the labels, and the ambiguity of the labels to position the labels in suitable order. They utilize Summed Area Table to evaluate the overlaps, conflicts, and ambiguity independently \revDr{on}\revDp{of} label size and screen resolution.

\subsection{Slider Models}
\label{subsec:sliding-models}
\revB{Compared to fixed-position models, slider models provide greater label placement flexibility by allowing continuous label movement around their corresponding features. However, slider models have not received much attention from the scientific community, and only a few approaches have been proposed.}
For instance, Hirsh~\cite{Hirsch1982} proposed a forced-based algorithm, where the translation vectors are repeatedly computed based on repelling forces for overlapping labels. 
Doddi~\etal~\cite{Doddi1997} introduced a slider model\revDr{,} where the goal is to maximize the size of square or circle-shaped labels.
The presented method allows rotating each label around the corresponding feature. 
Kreveld~\etal~\cite{MarcVanKreveld1999} studied several sliding models (top-, two- and four-slider model\revDg{s}) and proposed \revDg{a} combinatorial polynomial-time greedy approximation approach for all mentioned slider models. 
Li~\etal~\cite{Li2016} \revDr{proposed a method based on the region of movability, that comes from a plane collision detection theory. The idea is to define a complete conflict-free search space for label placement and then derive the best position for each label using heuristic search methods.}\revDg{proposed a method rooted in \revDp{the} plane collision detection theory, using the region of movability to define a conflict-free search space for label placement and determine the best label positions through heuristic search.} 
The mentioned approach of Luboschik~\etal~\cite{Luboschik2008} \revDr{is using the slider model as well. The approach tries to position each label sequentially with the fixed-position model, slider model, and \revA{finally as a distant label connected with the point feature by a straight-line leader}. For the distant labels, the positions of label candidates are sampled from a spiral function. The authors suggest that the proposed approach enables interactive labeling up to 1000 features.}\revDg{also utilizes the slider model, sequentially positioning labels using fixed-position, slider, and distant label models with straight-line leaders. For distant labels, they sample candidate positions from a spiral function. Their approach supports interactive labeling for up to 1000 features.}

\subsection{Machine Learning \& Visualization}
\label{subsec:ml4vis}

\revC{%
There has been a recent trend of incorporating machine learning techniques into visualization \revD{(ML4VIS)} to enhance the efficiency of visualization creation and suggest appropriate visual representations.}
\revC{%
Machine learning can be broadly categorized into several types, including supervised learning and reinforcement learning, each with its unique characteristics and suitable application areas. Supervised learning is the most common type, where an algorithm learns a model from labeled training data and then uses this model to make predictions for unseen data.}
\revC{%
In the context of visualization, supervised learning has been utilized in several impactful ways. For instance, Luo~\etal~\cite{Lu2019} introduced DeepEye, a system that leverages supervised learning to recommend suitable visualizations based on the provided data. Likewise, Bylinskii~\etal~\cite{Bylinskii2017} formulated a method for predicting users' visual attention distribution on infographics. Similarly, Chen~\etal~\cite{Chen2020} employed a series of supervised-learning techniques for the automatic segmentation of graphical elements from timeline infographics.
}

\revC{%
Reinforcement learning represents another significant category of machine learning, where an agent learns to make decisions by taking actions in an environment to maximize some notion of cumulative reward. Despite the success of reinforcement learning across various domains, its application in visualization \revD{(RL4VIS)} remains relatively unexplored, with only a handful of works adopting this technique. 
For instance, Tang~\etal~\cite{Tang2021} proposed PlotThread, which facilitates collaboration between \revDp{an} RL agent and \revDp{a} human designer to modify the storyline's layout. 
The trained agent assists designers and refines their interactions on a shared canvas.
Zhou~\etal~\cite{Zhou2021} proposed Table2Chart leveraging deep Q-learning and heuristic search to generate chart sequences from table data.
Hu~\etal~\cite{Hu2021} proposed a method for optimizing the coordinate ordering of sets of star glyphs related to multiple class labels to maximize perceptual class separation. 
In addition, Wu~\etal~\cite{Wu2021} developed MobileVisFixer, which automates a mobile-friendly visualization re-design process, and Deng~\etal~\cite{Deng2022} introduced DashBot, a Deep RL-based tool for generating analytical dashboards.
For a more comprehensive review of the use of machine learning in visualization, please refer to the survey by Wang~\etal~\cite{Wang2022}.
}%
\revD{While RL provides innovative solutions for visualization problems, visual analytics also proves instrumental in understanding and interpreting RL models (VIS4RL). For instance, DQNViz~\cite{Wang2018} proposed by Wang~\etal\ and DRLViz~\cite{Jaunet2020} introduced by Jaunet~\etal\ offer visual analytics approaches to understand deep Q-networks and deep reinforcement learning, respectively. Mishra~\etal ~\cite{Mishra2022} visualized explanations of agent behavior in reinforcement learning, whereas Wang~\etal~\cite{Wang2022b} employed visual analytics for RNN-based deep reinforcement learning.}

\revC{%
\revDr{Despite}\revDp{Although} the techniques above tackle various tasks, none of them address the specific challenge of label placement. Furthermore, RL-based techniques predominantly rely on a single-agent approach, a strategy that diverges from the multi-agent system proposed in our research.
Our work introduces a novel approach wherein multiple agents, acting as proxies for labels, are tasked to learn and interact concurrently within a shared environment. This paradigm inherently involves more complex dynamics, such as agent coordination and management of a non-stationary environment, which arise due to the simultaneous learning processes undertaken by the agents \cite{Lowe2017, Canese2021}. Therefore, our work represents not only an advancement in the application of reinforcement learning to the intricate problem of label placement but also \revDg{the} 
exploration of multi-agent systems in \revDp{the} visualization domain.
}

\section{\revC{Introduction to Reinforcement Learning}}
\label{sec:rl-introduction}

\revC{
\revBa{
Instead of relying on labeled data common for supervised learning, reinforcement learning (RL) leverages an agent that learns to take actions that maximize a cumulative reward by exploring the environment and observing the consequences of its actions. Reinforcement learning techniques have become an essential tool for solving various complex problems, particularly when dealing with decision-making. However, the curse of dimensionality in state space makes it challenging to apply RL effectively. To overcome this, RL is usually extended to Deep Reinforcement Learning (DRL), which employs a neural network as a function approximator~\cite{Sutton2020}. Based on the number of agents, RL can be further classified into a Single-Agent Reinforcement Learning (SARL) variant, where only one agent interacts with the environment. The second variant is Multi-Agent Reinforcement Learning (MARL), where multiple agents interact with a shared environment and work collaboratively to attain a shared goal.
}
}

\revC{
The formal basis of reinforcement learning is the theory of Markov Decision Processes (MDP). Over the discrete-time steps $t \in \mathbb{N}$, an agent observes the \textit{state} $s_t \in S$ of the \textit{environment}. Given the state, the agent selects \textit{actions} $a_t \in A$ based on a \textit{policy} $\pi$ that maps the state to a probability distribution over actions.
At the next time step $t+1$ as a consequence of \revDp{the} selected action, the agent receives feedback in a form of numerical \textit{reward} $r_{t+1}$ and transitions to new state $s_{t+1}$.\footnote{We follow the notation of Sutton and Barto~\cite{Sutton2020} and use the $r_{t+1}$ instead of $r_t$ to denote that reward and next state $s_{t+1}$ are determined jointly.} The sequence of states, actions, and rewards is called \textit{trajectory} or \textit{rollout}.
}%
\revC{%
The agent's goal is to find an \textit{optimal policy} $\pi_*$ that maximizes the \textit{discounted return} $G_t = \sum_{k=0}^{T}{\gamma^k r_{t+k+1}}$, where $\gamma \in [0, 1]$ is a \textit{discount factor}. As $\gamma \rightarrow 0$, the agent becomes more shortsighted and maximizes the immediate reward. On the other hand, the agent takes the future rewards into account more seriously when $\gamma \rightarrow 1$.
The agent's task is called \textit{episodic} if $T < \infty$, and such a trajectory is called an \textit{episode}.
The \textit{value function} $v_\pi(s)$, also called \revDp{the} state-value function, is the expected return beginning in state $s$ and following policy $\pi$ afterward. Formally, the value function is defined as $v_\pi(s) = \mathbb{E}_{\pi}[G_t | s_t = s]$.
Likewise, the \textit{q-value function} $q_\pi(s, a)$, also called action-value function, is the expected return beginning in state $s$, taking action $a$, and following policy $\pi$ afterward. Formally, the value function is defined as $q_\pi(s, a) = \mathbb{E}_{\pi}[G_t | s_t = s, a_t = a]$.
}%
\revC{%
In the DRL, the Markov property is often relaxed so that the agent does not have to be fully aware of the environmental state. Instead, the agent observes only a partial state known as \textit{observation} $o_t$ (retains far less information compared to the entire state of the environment), forming the Partially Observable MDP (POMDP). One step further is the cooperative Multi-Agent setting, where the agents act in parallel, extending the POMDP to the Decentralized POMDP (dec-POMDP). We refer to Sutton and Barto~\cite{Sutton2020} for further details on the formal background.
}

\section{Learning Point-feature Label Placement}
\label{sec:learning-pfl}

\revB{
Traditionally, supervised learning methods require a substantial amount of labeled data samples to train the model. However, this requirement poses a significant challenge, as no such labeled dataset currently exists for the point-feature label problem. Creating a dataset for this purpose would require access to an immense number of high-quality drawings with labels and time-consuming\revDr{ and expensive} annotations.
The novel approach presented in this work tackles this problem by posing the point-feature label placement as a reinforcement learning problem, which circumvents the challenge posed by the absence of ground-truth label placement datasets, a crucial requirement for traditional supervised learning methods.}

\revB{%
Designing the labeling problem as MARL has several significant benefits over SARL.
\revBa{%
First, when posing the label placement as SARL, the agent acts as a supervisor managing all labels at once. However, in that case, the observation and action space size changes with a varying number of labels, which contradicts the RL's prerequisite of fixed-sized observation and action spaces. On the contrary, with the abstraction of an agent for each label in MARL, an agent's individual observation and action\revDr{s} space can be implicitly designed as fixed-sized. Therefore, the number of agents (\ie copies of trained strategy) varies in MARL \revDr{instead of}\revDp{compared to} \revDp{the} observation and action space size in SARL.
}%
Second, even if one would overcome the variability, the SARL observation space would still be several times larger than in MARL, and the neural network architecture would be more complex and have significantly more parameters, resulting in more challenging training.
Third, one would have to collect many more trajectories to train the super agent, as the trajectory in SARL is a collection of the observations, actions, and rewards of all labels. On the contrary, the trajectory of each individual agent can be used to improve the shared strategy in MARL.
Therefore, due to \revDp{the} mentioned properties\revDr{ of SARL and MARL}, we decided to represent each label by an agent, which finally unfolds into \revDp{a} Multi-Agent Deep Reinforcement Learning (MADRL) problem.
From now on, we also refer to an agent as a label or a label agent interchangeably. Similarly, we refer to a point feature as an \textit{anchor}. 
}

The proposed method is designed explicitly for \textit{adjacent} PFL, meaning a label can be placed only around its anchor. Our intention is to find a conflict-free label position for each anchor (denoted as \textit{complete} labeling), and if such a position does not exist for all or cannot be found by the method, we call the labeling \textit{incomplete}. The emergence of label placement strategy and rules, in RL referred to as a policy, of our method is driven entirely by the learning process contrary to \revDp{the} existing hand-crafted methods.   %

\subsection{Environment}
\label{subsec:environment}

In the following text, we follow the terminology and definitions defined by Bekos~\etal~\cite{Bekos2019}\revDp{,} later extended by Bobák~\etal~\cite{Bobak2020}.

\begin{figure*}
    \subfigure[] {
        \label{fig:environment}
        \includegraphics[width=0.48\textwidth]{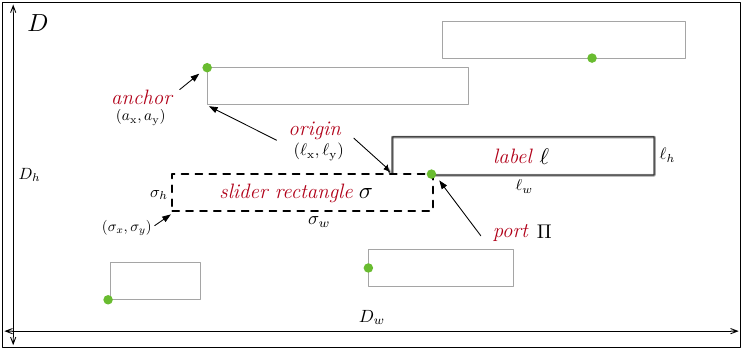}
    }
    \hfill
    \subfigure[] {
        \label{fig:observations}
        \includegraphics[width=0.48\textwidth]{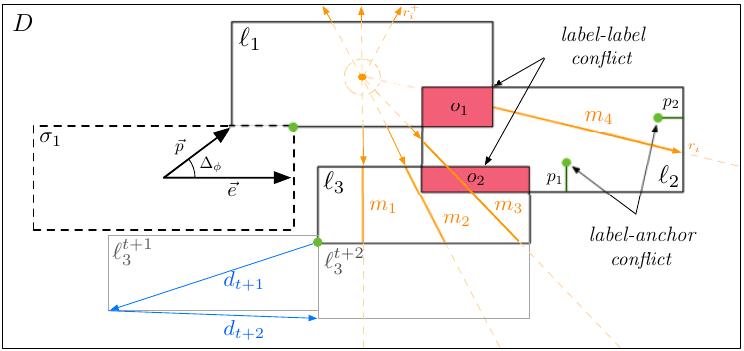}
    }
    \caption{\revB{Illustration \subref{fig:environment}} describes the used terminology and definitions. \revB{Illustration \subref{fig:observations}} depicts \revDp{the} observed modalities and defined action $\Delta_\phi$. \revDp{The} orange dashed lines represent rays providing the mapping vector of modalities. A ray begins at the label's center, but the actual reading of range starts at the label's bound. We collect the distance and type of the nearest intersection (\ie label, anchor, or bound) for each ray. The distance is positive (denoted by the plus sign subscript) if the ray hits the label from the outside. On the other hand, the distance is negative if the ray hits the label from the inside; see the ray denoted by $r_i^-$. Furthermore, we compute the mass of bodies of labels (denoted by the $m_{1\ldots4}$) that the ray went through till the bound of \revDg{the} environment. The overlap modality is depicted by the area of $o_1$ and $o_2$. The value for $\ell_1$ is just \revDg{the} area of $o_1$, but $o_1 + o_2$ for $\ell_2$ as it is in \revDp{a} label-label conflict with $\ell_2$ and $\ell_3$. Similarly, we demonstrate the penetration distance for $\ell_2$ as a sum of $p_1$ and $p_2$. Finally, we show the displacement distance $d_{t+1}$ and $d_{t+2}$ time steps $t+1$ and $t+2$, respectively. The total distance traveled is the cumulative sum of $d_t,\ t \in [0, T]$. \revA{The illustration does not show a complete description of the state for the illustration clarity (\eg only a few rays are visualized, \revDp{and the} labels corresponding to anchors producing the penetration $p_1$ and $p_2$ are not shown).}}
    \label{fig:definitions-env-obs}
\end{figure*}

\revC{Inspired by the interface of OpenAI Gym~\cite{OpenAIGym2016}, we transformed the adjacent PFL problem into a custom-developed environment, referred to as \texttt{AdjacentPFLEnv}, to facilitate the reinforcement learning paradigm.} The proposed environment consists of a set of anchors $\mathcal{A}$, each defined by its coordinates $(a_{\mathrm{x}}, a_{\mathrm{y}})$ enclosed within rectangular \textit{drawing region} $D$ of dimensions $(D_\mathrm{w}, D_\mathrm{h})$. Each anchor \revDr{has additional information attached in the form of}\revDg{is paired with} an axis-aligned box denoted as \textit{label agent} $\ell$ defined by its \textit{origin} coordinates $(\ell_{\mathrm{x}}, \ell_{\mathrm{y}})$ and dimensions $(\ell_\mathrm{w}, \ell_\mathrm{h})$. 
\revDg{A set of all label agents within the environment is denoted as $\mathcal{L}$.}
The label agent's origin coordinates lie on the circumference of the \textit{slider rectangle} $\sigma$, whose origin is defined as $(\sigma_x, \sigma_y) = (a_x-\ell_w, a_y-\ell_h)$ and dimensions as $(\sigma_w, \sigma_h) = (\ell_\mathrm{w}, \ell_\mathrm{h})$.
\revDg{Finally, each label agent $\ell$ remains tethered to its respective anchor via an attachment point denoted as \textit{port}~$\Pi$}\revC{, see \autoref{fig:environment}}.
\revB{We derive the initial origin of an associated label agent as
\begin{subequations}
\label{eq:label-origin}
\begin{align}
    \label{eq:label-origin-x}
    \ell_x &= \mathrm{clip(a_x, 0, D_w - \ell_w)}, \\
    \label{eq:label-origin-y}
    \ell_y &= \mathrm{clip(a_y, 0, D_\revA{h} - \ell_h)},
\end{align}
\end{subequations}
where $\mathrm{clip}(x, b_l, b_u)$ is a piecewise function that clips the value $x$ between lower $b_l$ and upper $b_u$ bound. Therefore, a label agent is placed at the initial state $s_0$ primarily at the most preferred position in the upper \revA{right} quadrant of the 4-position model, despite the fact that agents can be in \textit{conflict} (\ie agent-agent overlap, agent-anchor penetration), see \autoref{fig:observations}.
\revDr{A set of all label agents within the environment is denoted as $\mathcal{L}$. A label agent $\ell$ is connected to the corresponding anchor at an attachment point denoted as \textit{port}~$\Pi$.}}

\begin{figure*}
    \subfigure[] {
        \label{fig:delta_0}
        \includegraphics[scale=0.99, fbox]{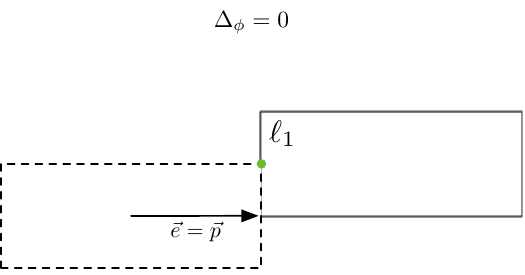}
    }
    \hfill
    \subfigure[] {
        \label{fig:delta_90}
        \includegraphics[scale=0.99, fbox]{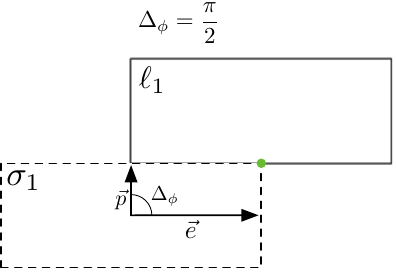}
    }
    \hfill
    \subfigure[] {
        \label{fig:delta_180}
        \includegraphics[scale=0.99, fbox]{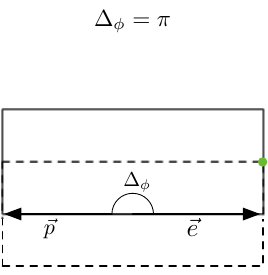}
    }
    \hfill
    \subfigure[] {
        \label{fig:delta_270}
        \includegraphics[scale=0.99, fbox]{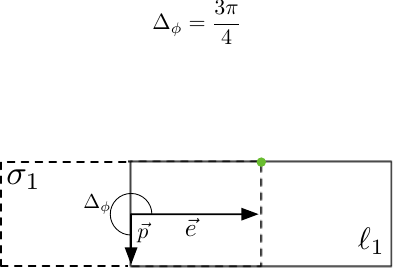}
    }
    \caption{\revB{A detailed illustration of \revDp{the} agent's action $\Delta_{\phi}$. For $\Delta_{\phi}: 0 \rightarrow 2\pi$, the origin of the label $[\ell_x,\ \ell_y]$ moves counterclockwise along the circumference of a slider rectangle $\sigma$ (denoted by a dashed line), whose origin is defined as $(\sigma_x, \sigma_y) = (a_x-\ell_w, a_y-\ell_h)$ and dimensions as $(\sigma_w, \sigma_h) = (\ell_\mathrm{w}, \ell_\mathrm{h})$.}}
    \label{fig:delta_action}
\end{figure*}

\revB{In the training phase, the anchor coordinates and dimensions of associated label agents are randomized at the initial state $s_0$. Coordinates of an anchor are drawn from the uniform distribution such as $a_x \sim U(0, D_w)$ and $a_y \sim U(0, D_h)$. Dimensions of label agents are determined in the similar fashion $\ell_w \sim U(0.1D_w, 0.15D_w)$, and we fixed $\ell_h = 0.05D_h$.}
The environment is populated \revBa{only} by one to two label agents, \revDr{which none or both}\revDp{none of which or both of which} overlap with the other agent, and we chose $D_w = 600$, $D_h = 400$ pixels. The environment terminates at the fixed horizon of $T=100$ steps. \revA{After termination, a new randomized configuration of label agents is populated in the environment and presented to the training algorithm. We apply a fixed-horizon approach to avoid a non-episodic behavior (\ie infinite horizon) when conflict-free label placement does not exist. Furthermore, a fixed horizon helps stabilize the label agent's position after finding a conflict-free arrangement by allowing the agent to discover that any additional action can lead to a deterioration of the reward.}

\revBa{In the evaluation phase, we can populate the environment with any number of label agents, even though we trained the policy with only one to two agents. We are leveraging the capability of RL to generalize to \textit{instances} (\ie specific configurations of an environment) unseen during the training to find the label placement for hundreds of anchors rather than just two. Moreover, the size of the drawing area, label size, and other mentioned parameters can be selected arbitrarily.}

\subsection{\revB{Observation Space}}
\label{subsec:observation-space}

\revB{%
We represent the state of the environment for each label agent $\ell$ solely
by \revDp{the} local same-shaped \textit{observation vector} $o_t^\ell$. This approach enables us to leverage all individual trajectories to train a shared policy and facilitate decentralized execution. 
\revC{Furthermore, we solely rely on sensor-based data, raw data acquired directly from the environment, instead of image data. Image data, represented as raster images or pixel matrices, typically demand greater storage and computational resources. By opting for sensor-based data, we effectively reduce the size of the observation vectors.} To capture the semantics of the observed modalities, we divided the observation $o_t^\ell$ into two distinct vectors: the \textit{mapping vector} $M$ and the \textit{self-aware vector} $S$. The observation vector $o_t^\ell$ is then obtained by concatenating $M$ and $S$.
}

\revB{%
The modalities captured by the mapping vector $M = [\texttt{d}, \texttt{t}, \texttt{c}, \texttt{m}]$ serve to encode the agent's surroundings through the use of 32 ray sensors, which are uniformly distributed around the label boundaries and function in a similar manner to LiDAR sensors. These sensors measure the distance \texttt{d} to the closest intersection point, as well as the type \texttt{t} of an object that the ray intersects (\ie label, anchor, bounds of the environment). Additionally, the mapping vector captures the count \texttt{c} and mass \texttt{m} of the labels that the ray passed through.
}

\revB{%
The self-aware vector $S = [\texttt{O}, \texttt{D}, \texttt{Ape}, \texttt{Apr}, \texttt{Ad}, \texttt{T}]$ supplies local modalities that pertain mainly to the agent's conflicts, including overlaps with other agents and penetrations with anchors. We define the overlap modality \texttt{O} as a sum of the occluded area between the given agent and the other agents being in conflict. The displacement \texttt{D} represents the \revC{E}uclidean distance of the agent's origins between two consecutive time steps $t$ and $t+1$.  
Furthermore, we define penetration distance as the \revC{E}uclidean distance between the penetrated anchor and the nearest point of escape on the circumference of the label. Like the overlap modality, we define the penetration modality \texttt{Ape} as a sum of the penetration distances between a given agent and its anchor or the other agents' anchors being in conflict. For both modalities, we also provide the agent \revDp{with} a count of conflicts relative to the total number of label agents.
Additionally, the agent observes the \revC{E}uclidean distance to its anchor \texttt{Ad}, \revDp{the} relative position of \revDp{the} anchor to \revDp{the} label agent's port \texttt{Apr}, and finally, the information about elapsed time steps \texttt{T}.
}

\revB{%
\revA{%
We normalize all observation modalities. An overview of the observed modalities is presented in \autoref{fig:observations}, while a comprehensive description is available in the supplementary material\footnote{\label{foot:supplement}\revC{Supplementary material related to this project can be accessed via our project page at \url{http://cphoto.fit.vutbr.cz/reinforced-labels}.}}.}
}

\subsection{Action Space}
\label{subsec:action-space}

\revA{We designed the action space of agents such that they can change the origin continuously without discretization to the environment's raster, allowing subpixel precision and independence of the drawing area dimensions.}
Therefore, we define a continuous action $\Delta_\phi$ representing the angle between basis vector \revB{$\vec{e}=(1, 0)$} and vector $\vec{p}=(p_x, p_y)$, where point $[p_x, p_y] \equiv [\ell_x,\ \ell_y]$ belongs to the circumference of the slider rectangle for a given label $\ell$, see \autoref{fig:observations} and \autoref{fig:delta_action}. 
\revC{We use one-dimensional action space $\Delta_\phi$ rather than an intuitive two-dimensional action space of $\Delta_x$ and $\Delta_y$ to simplify the learning process. In our experience, multi-dimensional action can introduce unnecessary coordination complexity.}
Given $\Delta_\phi \in (0, 2\pi)$, we define the origin of a label in\revDr{ the} quadrant Q1 as
\begin{equation}
    \label{eq:action}
    [\ell_x,\ \ell_y]_{t+1} =
    \begin{cases}
    \frac{\ell_w}{2}[1,\ \tan(\Delta_\phi)] & 0 \leq \Delta_\phi < \arctan\left(\frac{\ell_h}{\ell_w}\right) \\
    \frac{\ell_h}{2}[\cot(\Delta_\phi),\ 1] & \arctan\left(\frac{\ell_h}{\ell_w}\right) \leq \Delta_\phi < \frac{\pi}{2}.
    \end{cases}
\end{equation}
The derivation for\revDr{ other} quadrants \revDr{Q2, Q3, and Q4}\revDg{Q2 to Q4} is trivial -- the idea is to transform the $\Delta_\phi$ into the Q1 and then flip the position to the corresponding quadrant. For \revDr{simplicity of previous explanation}\revDg{clarity}, we omitted \revDr{that we employ the definition}\revDg{use} of $\Delta_\phi \in [-1, 1]$ in our implementation; nevertheless, the previous declaration holds. 

\revA{%
To facilitate continuous action space with an infinite number of actions, we define a policy as a parameterized Gaussian distribution. Therefore, instead of learning the probabilities of all possible actions, which is infeasible, we learn statistics of the distribution. Formally, we define policy as}
\begin{equation}
    \label{eq:agent-policy}
    \pi_\theta(\Delta_\phi | s_t) = \mathcal{N}\left(\mu_{\theta}(s_t), \sigma_{\theta}^2(s_t)\right),
\end{equation}
where $s_t$ is the current state at time step $t$, $\mu_{\theta}(s_t)$ and $\sigma_{\theta}^2(s_t)$ is the mean and variance of the distribution parametrized by the neural network parameters $\theta \in \mathbb{R}^d$ further described in \autoref{subsec:policy-value-nn}.

\subsection{Reward}
\label{subsec:reward}

Our goal in this work involves the emergence of cooperation among agents to find a conflict-free label position for each anchor.
Therefore, we combine two types of rewards as suggested by Nguyen~\etal~\cite{Nguyen2018}. The \textit{local} reward $r^{\mathrm{local}}$ assigns an agent its feedback solely based on individual efforts, \revDr{and contrary,}\revDp{while} the \textit{global} reward $r^{\mathrm{global}}$ gives an agent its feedback \revDr{founded}\revDp{based} on the entire state of the environment.

As a local reward, the agent receives step-wise penalization (\ie negative reward) for being in label-label conflict with another agent, meaning both label agents overlap with each other. We compute the \textit{overlap value} $o(\ell)$ \revDp{in the} same \revDp{way} as the overlap modality, as a sum of the occluded area between a given agent and the other agents in conflict.

The global reward is the composition of local rewards among individual agents. Formally, we define the local reward $r^{\mathrm{local}}$ and global reward $r^{\mathrm{global}}$ as
\begin{align}
    \label{eq:reward}
    r_{t+1}^{\mathrm{local}}(\ell) = -o(\ell) && %
    r_{t+1}^{\mathrm{global}} = \sum_{\ell \in \mathcal{L}} r_{t+1}^{\mathrm{local}}(\ell),
\end{align}
where $t$ is \revDp{the} current time step, $o(\ell)$ is the overlap value.
Based on these definitions, we define the total reward for a label agent $\ell$ as

\begin{equation}
    \label{eq:total-rew}
    r_{t+1}^{\mathrm{total}}(\ell) = (1-w) \cdot r_{t+1}^{\mathrm{global}} + w \cdot r_{t+1}^{\mathrm{local}}(\ell). 
\end{equation}
By observing the influence of various weight values on the final reward of the trained policy, we recommend using $w=0.5$.

\subsection{Policy \& Value Network Architecture}
\label{subsec:policy-value-nn}

\begin{figure*}
    \includegraphics[width=\textwidth]{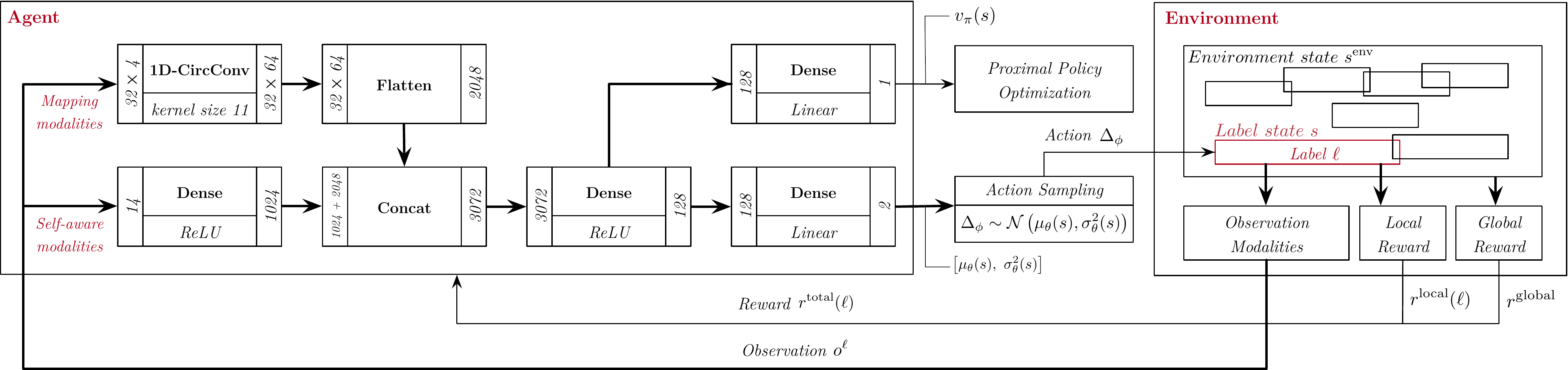}
    \caption{Overview of Reinforced Labels and its neural network architecture. The proposed \texttt{AdjacentPFLEnv} environment describes the adjacent point-feature labeling problem. We are proposing a Multi-Agent approach, where an agent controls the origin of a label. We use parameter sharing, meaning each agent acts according to the same policy network with identical parameters. Therefore, we can use trajectories of individual agents to optimize the parameters of the proposed network. We utilize only local agent observations \revBa{$o^\ell$}, enabling parallel execution in the evaluation phase. We provide the agent feedback by local rewards $r^{\mathrm{local}}$ and global rewards $r^{\mathrm{global}}$ only during the training; the reward is not used in the evaluation phase. The mapping and self-aware modalities are passed through the network, resulting in a state value $v(s)$ and parameters of normal distribution $\mu_\theta(s)$ and $\sigma_\theta^2(s)$. The training algorithm~\cite{Schulman2017} uses the $v(s)$ to optimize the parameters of the network. The distribution parameters are used to sample an action $\Delta_\phi$. Finally, the agent acts according to the action, which might translate (\ie the $\Delta_\phi \neq 0$) \revDp{in}to the change of the label's origin.}
    \label{fig:rl-scheme}
\end{figure*}

We designed an efficient yet straightforward feedforward network with just less than half of a million (\revA{412} thousand) parameters, as depicted in \autoref{fig:rl-scheme}. The architecture consists of two input heads -- \textit{mapping} and \textit{self-aware}, and two output branches -- \textit{value} and \textit{policy}.
\revA{The ray observations from sensors constituting the mapping head are first passed through a circular 1D-convolution layer proposed by Schubert~\etal~\cite{Schubert2019} to capture the correlation between individual readings, including the borders of the tensor. The intermediate representation is reshaped to form a 1D tensor.
The other observation modalities forming the self-aware head are concatenated and embedded by a dense layer. 
The mapping and self-aware heads are concatenated and passed by a final shared dense layer. At the end of the architecture, two separate dense layers split the outcome into \revDp{the} value and policy branches.
}

\subsection{Training}
\label{subsec:training}
\revA{%
\revC{We adopt a \textit{parameter sharing} in which each agent utilizes the same policy network with identical parameters. This approach allows us to optimize the parameters of the proposed network using the trajectories of \revDp{the} individual agents.}
We update the parameters of the policy network in \revC{a} \textit{policy gradient} fashion -- meaning the parameters are updated based on \revC{the} gradient of an estimate of expected return with respect to the policy parameters. In particular, we utilize Proximal Policy Optimization (PPO), one of the most prominent actor-critic policy gradient methods proposed by Schulman~\etal~\cite{Schulman2017}. The method is best known for its relative simplicity while preserving the convergence properties of more complex predecessors.
We refer to the survey of Arulkumaran~\etal~\cite{Arulkumaran2017b} for further details on the DRL algorithms.
The goal is to optimize the parameters $\theta$ that maximize the expected discounted return $\mathbb{E}[G_t]$. The surrogate objective of PPO is defined as 
\begin{equation}
\label{eq:ppo-clip}
    L(\theta)^{CLIP} = \mathbb{E}\left[\min(r_t(\theta) \hat{A}_t, \mathrm{clip}(r_t(\theta), 1-\epsilon, 1+\epsilon)\hat{A}_t)\right],
\end{equation}
where $r_t(\theta) = \frac{\pi_\theta(a_t | s_t)}{\pi_{\theta{old}}(a_t | s_t)}$ is the probability ratio of the new and old polic\revDp{ies}. Value of $\hat{A}_t$ is the estimator of the \textit{advantage function}, computed by Generic Advantage Estimation introduced by Schulman~\etal~\cite{Schulman2016a}, describing whether choosing action $a_t$ in state $s_t$ is better or worse than the average action of policy~$\pi$. 
One of the essential concepts of PPO is the clip operation on top of the probability ratio $r_t(\theta)$ that discourages the policy from dramatically changing between training iterations, resulting in convergence issues. Lower values of $\epsilon$  correspond to more consistent policy improvements. On the other hand, higher values yield greater variance and volatility of convergence.} 
\revA{%
The value network objective is formulated as regression via mean square error as
\begin{equation}
L^{VF}(\theta) = \mathbb{E}\left[(v_\pi(s_t | \theta) - v^\mathrm{target})^2\right],
\end{equation}
where $v^\mathrm{target} = r_t + \gamma r_{t+1} + \ldots + \gamma^{T-t} v_{\pi_{\theta{old}}}(s_t)$.
Because the proposed neural network architecture shares parameters between policy and value networks, we combine \revDr{previously said}\revDp{the aforementioned} objectives to a composed loss to train both networks simultaneously as $L(\theta) = L(\theta)^{CLIP} - L^{VF}(\theta)$. A detailed list of hyperparameters is available in the supplementary material.
}

\subsection{Inference}
\label{subsec:inference}
\revA{Once the policy network is trained, we can use it to evaluate an instance with an arbitrary number of anchors, size of labels, and dimensions of the drawing area, all due to the proposed MADRL design properties and choice of the observed modalities and actions.} At the beginning of the evaluation, we initialize the environment according to  \autoref{eq:label-origin}. The evaluation step for an agent $\ell$ consists of a collection of observation modalities \revB{to form vector $o_t^\ell$} and passing it through the \revC{shared} policy network, resulting in\revDr{the} action~$\Delta_\phi$. This sequence produces a new environment state $s_{t+1}$ and a new observation vector $o_{t+1}^\ell$. The process is repeated for all agents until no conflicts are present or the horizon $T$ is hit.

\revC{\section{Ablation Study}}
\label{sec:ablation-study}
\revC{
We conducted an ablation study to fine-tune the proposed method and validate our design choices. The goal is to shed light on the impact of each component within our method and to understand how variations in these components can influence overall performance. Additionally, the findings from these ablations open up the potential for future research pathways, creating opportunities for further optimization and refinement of our technique. To measure the performance of our method under different ablations, we introduce a novel metric and benchmark dataset.}

\revC{
\subsection{Completeness Metric}
\label{subsec:completeness-metric}
}

\revD{Labeling metrics typically measure the quantity of non-conflicting labels. Such metrics, however, are best suited for methods that ensure conflict-free label layouts. Conversely, methods not offering such assurances may calculate performance based on the sum of overlapping regions. Yet, from the perspective of \revDp{the} established label placement rules~\cite{Yoeli1972, Imhof1975}, any overlap, whether slight or significant, deems a layout non-conflict-free. Given this landscape, we observed a gap: while conflict-free methods may omit labels to avoid conflicts, it is uncertain how the overlap regions would appear. Conversely, methods without a conflict-free guarantee do not indicate labels that should be discarded to achieve a conflict-free layout. To close the gap, we introduce the \textit{completeness} metric.}

Our work focuses on finding a conflict-free label position
for each anchor. If such a position does not exist for all anchors or cannot be found by the method, we call the labeling \textit{incomplete}. On the other hand, we denote the labeling as \textit{complete} if all anchors are annotated without conflict.

We measure the performance of label placement methods by the completeness metric representing the percentage of complete labelings for a given set of instances. For example, let D be a dataset of 10 instances. Let $M_1$ be 
PFL method that found eight complete conflict-free layouts without the need of removing any label and two incomplete conflict-free layouts with several removed labels. Let $M_2$ be a method with the same PFL properties as $M_1$ that found nine complete conflict-free layouts and one incomplete layout with remaining conflicts (\ie label-label or label-anchor conflict). The completeness of $M_1$ is 80\%, and the completeness of $M_2$ is 90\%, as the latter method found more complete layouts out of a set of ten given instances. Therefore,\revDr{ the} method $M_2$ performs better than\revDr{ the} method $M_1$.

\revC{\subsection{Benchmark Dataset}
\label{subsec:benchmark-dataset}}
\revA{%
We have created a benchmark dataset to compare outcomes and performance among the evaluated label placement methods. We split the dataset into two parts -- \textit{compact} and \textit{volume} datasets.}
\revA{%
To endorse the standardized evaluation of labeling methods, we provide the benchmark dataset as supplementary material.}

\revA{%
We generated the anchor coordinates using a pseudo-random number generator with uniform distribution \revB{common across multiple bodies of previous work \cite{Pavlovec2022, Li2016, Yamamoto2002b}}. In the compact dataset, we sampled the anchor coordinates from an area of 600$\times$400 and sequentially increased the number of drawn samples by five, ranging from 5 to 50. For the volume dataset, we drew coordinates from an area of 2400$\times$1600 and consecutively raised their count by 50, going from 100 to 600. For each number of anchors, we generated ten instances. Therefore, the entire dataset consists of $41 \;250$ anchors divided into 210 instances. Furthermore, we randomly formed the corresponding labels so that the text consists of \revC{three} to \revC{seven} capital letters from the English alphabet.}

\revA{%
It is worth noting that the benchmark dataset also contains instances that cannot be solved, as it was created using a pseudo-random number generator, due to the factual inexistence of a conflict-free layout. In such a case, any labeling method cannot produce complete labeling. However, such instances do not influence the comparison of labeling methods on the proposed dataset because these instances always affect all\revDr{ labeling} methods the same.}

\subsection{Architecture Ablations}
\label{subsec:arch-ablations}

\begin{table}[t!]
\centering
\caption{Ablations of neural network architecture evaluated on the compact dataset. We computed the overall completeness metric for the dataset to summarize the performance and \revDr{ease comparing examined ablations}\revDp{simplify the comparison of the ablations examined}. Furthermore, we provide the average number of elapsed steps needed to solve instances in the dataset, inference time per step, and \revDp{the} number of parameters to illustrate the complexity of the neural network. \revC{The notation is explained  in \autoref{subsec:arch-ablations}}}.
\label{tab:arch-ablations}
\setlength{\tabcolsep}{10pt}
\begin{adjustbox}{width=\columnwidth}
\begin{tabu}{RlBcBcBcBcBc}
    \toprule
     Model & \textit{Completeness} & \textit{Steps} & \textit{Inference} & \textit{Parameters} \\ 
     & \% & - & ms & millions \\
     \midrule
     \rowstyle{\bfseries}\texttt{RFL\_Conv} & 97.9 & 30.721 & 3.077 & 0.412  \\
     \texttt{RFL\_ConvLn} & 96.8 & 37.333 & 3.513 & 0.412 \\
     \midrule
     \rowstyle{\bfseries}\texttt{RFL\_2Dns} & 96.4 & 41.466 & 2.775 & 0.673 \\
     \texttt{RFL\_2DnsLn} & 94.5 & 59.353 & 2.998 & 0.673 \\
     \midrule
     \rowstyle{\bfseries}\texttt{RFL\_1Dns} & 95.4 & 53.081 & 2.665 & 0.833 \\
     \texttt{RFL\_1DnsLn} & 94.7 & 63.307 & 2.882 & 0.833 \\
     \midrule
     \texttt{RFL\_rllib} & 90.6 & 81.905 & 2.008 & 0.206 \\
    \bottomrule
\end{tabu}
\end{adjustbox}
\end{table}

\revC{\revA{%
In order to justify the design choices behind our proposed architecture, we performed several architectural ablations. Each ablation was evaluated on the compact dataset ten times to adequately account for the inherent stochastic properties of Reinforced Labels (\texttt{RFL}). In an effort to summarize the performance and facilitate the comparison of the examined ablations, we report the overall completeness metric as a single average value for the compact dataset. The results of these evaluations can be found in \autoref{tab:arch-ablations}.}}

\revA{%
We categorized the ablations into four groups, starting with the proposed architecture and systematically removing/varying its components. Furthermore, we also evaluated the universal baseline architecture of \revBa{a leading reinforcement learning library RLLib utilized in our development~\cite {liang2018rllib}.}
The first ablation group \revC{\textbf{\texttt{RFL\_Conv*}}} contains a custom two-head, two-branch convolution-based architecture as described in \autoref{subsec:policy-value-nn}. The mapping modalities are first concatenated \revDr{to form}\revDg{into} vector $M$, embedded by a circular 1D-convolution layer, and optionally passed through a layer normalization. The self-aware modalities are concatenated \revDr{to form}\revDg{into} vector $S$, embedded by a dense layer, and optionally passed through a layer normalization \revC{denoted by \textbf{\texttt{Ln}} suffix}. Ultimately, the outcomes are concatenated, and two separate dense layers split the outcome into \revDp{the} value and policy branches.
The second ablation group \revC{\textbf{\texttt{RFL\_2Dns*}}} \revDr{contains custom}\revDg{retains} the two-head, two-branch architecture \revDr{without}\revDg{but excludes} the circular 1D-convolution layer. \revDr{The architecture is the same as the \texttt{RFL\_Conv*} except for}\revDg{Instead,} the mapping modalities are embedded by a dense layer.
The third ablation group \revC{\textbf{\texttt{RFL\_1Dns*}}} includes custom single-head, two-branch architecture. \revB{\revDp{The} observed modalities $M$ and $S$ are merged}, embedded by two dense layers, and optionally passed through a layer normalization. At the end of the architecture, two separate dense layers split the outcome into \revDp{the} value and policy branches.
The final ablation \revC{\textbf{\texttt{RFL\_rllib}}} holds the default baseline architecture of RLLib. \revB{\revDp{The} observed modalities $M$ and $S$ are concatenated} and separated into \revDp{the} value and policy by two triple-dense-layer branches.
}

\revA{%
\autoref{tab:arch-ablations} shows the results of \revDp{the} architectural ablations. The best\revD{-}performing \revD{model} is the \texttt{RFL\_Conv}, as described \revDg{in} \autoref{subsec:policy-value-nn}, achieving \revDg{an} \revB{overall} completeness of 97.9\%. We emphasize the importance of the circular 1D-convolution layer as, in addition to classical convolution, the circular one captures the correlation \revDr{on}\revDg{between} the borders of the tensor. The second-best score \revDr{earned}\revDp{was achieved by} \texttt{RFL\_2Dns}, without circular convolution but preserving the two-head scheme, achieving 96.4\%. 
However, the minor decline in completeness is accompanied by a significant increase in the average number of steps needed to solve instances in the dataset.
Changing the architecture to a single-head scheme (\texttt{RFL\_1Dns}) leads to a further decrease in performance to the completeness of 95.4\%. We also evaluated the default universal model \texttt{RFL\_rllib}, which performs the worst, achieving significantly lower completeness of 90.6\%.
The main difference between the architecture of \texttt{RFL\_rllib} \revDr{to}\revDp{and the} other variants is the absence of a shared dense layer before splitting the outcomes into \revDp{the} value and policy branches. Therefore, we argue \revDg{that} the presence of a shared dense layer is \revDg{a} vital part of the proposed architecture as it carries the most significant difference in completeness.
Finally, the results of our ablation study consistently show that layer normalization is an inappropriate architecture component for \revDp{the} given label-related modalities, always resulting in an overall performance drop. 
In the following text, we simplify our notation and use \texttt{RFL} to denote the \texttt{RFL\_Conv} model.
}

\subsection{Observation Ablations}
\label{subsec:obs-ablations}

\begin{table}[t!]
\centering
\caption{Ablations of observed modalities evaluated on the compact dataset. We provide overall completeness, the average number of elapsed steps needed to solve the dataset, inference time per step, and observation creation time per step. \revC{The notation is explained  in \autoref{subsec:obs-ablations}}}
\label{tab:obs-ablations}
\setlength{\tabcolsep}{2pt}
\begin{adjustbox}{width=\columnwidth}
\begin{tabu}{RlBcBcBcBcBc}
    \toprule
     Set & Modalities & \textit{Completeness} & \textit{Steps} & \textit{Inference} & \textit{Observation}  \\ 
     & & \% & - & ms & ms \\
     \midrule
     \nextrow\label{row:M008[dcm]S[OApeAprAdT]} & \texttt{M008[d cm]S[O ApeAprAdT]} & 96.8 & 36.791 & 2.723 & 19.870 \\
     \nextrow\label{row:M016[dcm]S[OApeAprAdT]} & \texttt{M016[d cm]S[O ApeAprAdT]} & 96.8 & 41.389 & 2.713 & 23.444 \\
     \rowstyle{\bfseries}\nextrow\label{row:M032[dcm]S[OApeAprAdT]} & \texttt{M032[d cm]S[O ApeAprAdT]} & 97.9 & 30.721 & 2.883 & 30.645 \\
     \nextrow\label{row:M064[dcm]S[OApeAprAdT]} & \texttt{M064[d cm]S[O ApeAprAdT]} & 97.9 & 32.159 & 3.077 & 40.400 \\
     \nextrow\label{row:M128[dcm]S[OApeAprAdT]} & \texttt{M128[d cm]S[O ApeAprAdT]} & 97.7 & 36.630 & 3.150 & 63.260 \\
     \midrule
     \rowstyle{\bfseries}\nextrow\label{row:M[dcm]S[OApeAprAdT]} & \texttt{M[d cm]S[O ApeAprAdT]} & 97.9 & 30.721 & 3.077 & 30.645 \\
     \nextrow\label{row:M[OrSi]S[OApeAprAdT]} & \texttt{M[OrSi]S[O ApeAprAdT]} & 96.8 & 46.382 & 2.679 & 16.013 \\
     \nextrow\label{row:M[dcm]S[ApeAprAdT]} & \texttt{M[d cm]S[ \ ApeAprAdT]} & 45.3 & 298.759 & 1.999 & 27.096 \\
     \nextrow\label{row:M[dcm]S[OAprAdT]} & \texttt{M[d cm]S[O \ \ \ AprAdT]} & 90.3 & 75.955 & 2.506 & 29.012 \\
     \rowstyle{\bfseries}\nextrow\label{row:M[d cm]S[OApeAdT]} & \texttt{M[d cm]S[O Ape \ \ AdT]} & 97.9 & 33.594 & 2.575 & 26.721 \\
     \nextrow\label{row:M[dcm]S[OApeAprT]} & \texttt{M[d cm]S[O ApeApr \ T]} & 97.4 & 34.255 & 2.698 & 28.690 \\
     \nextrow\label{row:M[dcm]S[OApeAprAd]} & \texttt{M[d cm]S[O ApeAprAd ]} & 97.3 & 33.095 & 2.722 & 28.606 \\
     \nextrow\label{row:M[cm]S[OApeAprAdT]} & \texttt{M[ \ cm]S[O ApeAprAdT]} & 96.7 & 36.610 & 2.569 & 27.742 \\
     \nextrow\label{row:M[dm]S[OApeAprAdT]} & \texttt{M[d \ m]S[O ApeAprAdT]} & 97.0 & 41.403 & 2.655 & 30.215 \\
     \nextrow\label{row:M[dc]S[OApeAprAdT]} & \texttt{M[d c ]S[O ApeAprAdT]} & 97.2 & 34.945 & 2.752 & 28.332 \\
     \midrule
     \nextrow\label{row:M[dtcm]S[ODApeAprAdT]} & \texttt{M[dtcm]\,S[ODApeAprAdT]} & 97.2 & 35.352 & 2.740 & 29.763 \\
     \nextrow\label{row:M[OrSi]S[ODApeAprAdT]} & \texttt{M[OrSi]\,S[ODApeAprAdT]} & 94.8 & 61.165 & 2.539 & 15.829 \\
     \nextrow\label{row:M[dtcm]S[DApeAprAdT]} & \texttt{M[dtcm]S[ DApeAprAdT]} & 50.6 & 270.392 & 2.007 & 41.181 \\
     \nextrow\label{row:M[dtcm]S[OApeAprAdT]} & \texttt{M[dtcm]S[O ApeAprAdT]} & 97.5 & 40.649 & 2.758 & 28.917 \\
     \nextrow\label{row:M[dtcm]S[ODAprAdT]} & \texttt{M[dtcm]S[OD \ \ AprAdT]} & 86.4 & 98.084 & 2.567 & 28.735 \\
     \nextrow\label{row:M[dtcm]S[ODApeAdT]} & \texttt{M[dtcm]S[ODApe \ \ AdT]} & 96.8 & 46.255 & 2.681 & 27.909 \\
     \nextrow\label{row:M[dtcm]S[ODApeAprT]} & \texttt{M[dtcm]S[ODApeApr \ T]} & 93.9 & 66.275 & 2.578 & 28.057 \\
     \nextrow\label{row:M[dtcm]S[ODApeAprAd]} & \texttt{M[dtcm]S[ODApeAprAd ]} & 95.1 & 55.721 & 2.609 & 27.961 \\
     \nextrow\label{row:M[tcm]S[ODApeAprAdT]} & \texttt{M[ tcm]S[ODApeAprAdT]} & 95.2 & 59.602 & 2.535 & 27.942 \\
     \rowstyle{\bfseries}\nextrow\label{row:M[dcm]S[ODApeAprAdT]} & \texttt{M[d cm]S[ODApeAprAdT]} & 97.6 & 41.915 & 2.912 & 30.325 \\
     \nextrow\label{row:M[dtm]S[ODApeAprAdT]} & \texttt{M[dt m]S[ODApeAprAdT]} & 96.4 & 54.344 & 2.579 & 28.562 \\
     \nextrow\label{row:M[dtc]S[ODApeAprAdT]} & \texttt{M[dtc ]S[ODApeAprAdT]} & 96.8 & 39.204 & 2.615 & 28.272 \\
     \midrule
     \rowstyle{\bfseries}\nextrow\label{row:M[dtcm]S[O]} & \texttt{M[dtcm]\,S[O \ \ \ \ \ \ \ \ \ ]} & 86.0 & 94.781 & 2.628 & 27.442 \\
     \nextrow\label{row:M[OrSi]S[O]} & \texttt{M[OrSi]\,S[O \ \ \ \ \ \ \ \ \ ]} & 83.3 & 96.557 & 2.522 & 14.972 \\
     \nextrow\label{row:S[O]} & \texttt{M[\ \ \ \ ]\,S[O \ \ \ \ \ \ \ \ \ ]} & 82.4 & 101.750 & 2.414 & 14.525 \\
     \bottomrule
\end{tabu}
\end{adjustbox}
\end{table}

\revA{%
Similarly to the previous section, we \revDr{evaluated ablations of}\revDg{ablated} \revDp{the} observed modalities to justify our design choices of representing the agent's state.
We start with the proposed observation set and then systematically remove/vary \revDp{the} included modalities fed into the best-performing architecture \texttt{RFL\_Conv}.
\revDr{To recall}\revDp{As described in \autoref{subsec:observation-space}}, the proposed vector of observed modalities $o_t$ consists of mapping modalities denoted by \revC{\textbf{\texttt{M}}} and self-aware modalities denoted by \revC{\textbf{\texttt{S}}}. Ray-based mapping modalities include distance to the nearest intersection \textbf{\texttt{d}}, type of the nearest intersected object \revC{\textbf{\texttt{t}}} (i.e., label, anchor, bounds of the environment), and count \revC{\textbf{\texttt{c}}} and\revDr{the} mass \revC{\textbf{\texttt{m}}} of labels that the ray went through. Self-aware observations include overlap \revC{\textbf{\texttt{O}}}, displacement \revC{\textbf{\texttt{D}}}, anchor penetration distance \revC{\textbf{\texttt{Ape}}}, anchor-port distance \revC{\textbf{\texttt{Apr}}}, anchor-origin distance \revC{\textbf{\texttt{Ad}}}, and time step \revC{\textbf{\texttt{T}}} modalities. To investigate the importance of ray-based modalities, we also include mapping modalities based on the agent's origin \revC{\textbf{\texttt{Or}}} and size \revC{\textbf{\texttt{Si}}} as a replacement \revDr{to}\revDp{of} ray casting.
}

\revA{%
The results are presented in \autoref{tab:obs-ablations}. For clarity, we provide a numerical notation of each set of evaluated modalities (\ie numbers in the first column of the table).
First and foremost, the results show that removing the overlap modality (comparing sets \ref{row:M[dtcm]S[DApeAprAdT]} with \ref{row:M[dtcm]S[ODApeAprAdT]} and \ref{row:M[dcm]S[ApeAprAdT]} with \ref{row:M[dcm]S[OApeAprAdT]}) dramatically degrades the performance and highly impacts the average number of steps needed to solve the labeling. Therefore, we argue that the overlap modality is the most \revDr{important}\revDg{crucial} modality that effectively allows an agent to avoid label-label conflicts.
Similarly, removing the anchor penetration modality \texttt{Ape} (comparing sets  \ref{row:M[dtcm]S[ODAprAdT]} with \ref{row:M[dtcm]S[ODApeAprAdT]} and \ref{row:M[dcm]S[OAprAdT]} with \ref{row:M[dcm]S[OApeAprAdT]}) leads to a significant performance decline and moderately impacts the average number of steps needed to solve the labeling. We argue that the penetration modality effectively allows an agent to avoid label-anchor conflicts.
Replacing the mapping ray-based modalities with modalities based on the agent's origin \texttt{Or} and size \texttt{Si} (comparing sets \ref{row:M[OrSi]S[ODApeAprAdT]} with \ref{row:M[dtcm]S[ODApeAprAdT]}) advocates the presence of ray casting in our method. Ray-based modalities deliver (a) higher completeness and (b) effectively cut the average number of needed steps nearly to half.
Interestingly, further \revDr{removing}\revDp{removal} of displacement modality \texttt{D} (comparing set \ref{row:M[dtcm]S[OApeAprAdT]} with \ref{row:M[dtcm]S[ODApeAprAdT]}) positively impacts the performance. We argue the displacement modality (cumulative distance traveled) decreases the number of steps needed for an expense of lower completeness. A similar explanation applies to ray-type modality (comparing set \ref{row:M[dcm]S[ODApeAprAdT]} with \ref{row:M[dtcm]S[ODApeAprAdT]}).
Therefore, we omitted displacement and ray-type modality due to the negative impact on completeness and conducted a second round of ablations (set \ref{row:M[dc]S[OApeAprAdT]} to \ref{row:M[dcm]S[OApeAprAdT]}). The results show an additional increase in completeness and a decline in the number of steps needed, \revC{please compare} set \ref{row:M[dtcm]S[ODApeAprAdT]} \revDp{with} \ref{row:M[dcm]S[OApeAprAdT]}. Further findings remain consistent with the first round of ablations (set \ref{row:M[dtc]S[ODApeAprAdT]} to \ref{row:M[dtcm]S[ODApeAprAdT]}) except for omitting the proximity modality (set \ref{row:M[d cm]S[OApeAdT]}) that achieves \revDp{the} same performance, but \revDp{a} higher number of steps \revDp{is} needed in comparison with set \ref{row:M[dcm]S[OApeAprAdT]}.}

\revA{%
Finally, we experimented with a number of rays, as shown by sets \ref{row:M008[dcm]S[OApeAprAdT]} \revDr{till}\revDp{through} \ref{row:M128[dcm]S[OApeAprAdT]}, where \revC{\textbf{\revD{the} number \revD{following} \texttt{M}}} denotes \revDp{the} number of casted rays. 
The results indicate that the best option is to cast 32 and 64 rays to achieve the best completeness. However, casting 64 rays demands more resources, leading to increased observation creation time compared to 32 rays. Reducing as well as increasing the number of rays leads to lower completeness. We \revDr{argue}\revD{hypothesize} that the information is likely too sparse at the lower end\revDr{ and does not provide enough information}\revD{, not providing sufficient detail}. Conversely, \revDr{a high number of}\revD{opting for 128} \revD{over 64 rays} \revDr{probably does not deliver}\revD{might not offer} \revD{substantial} additional information and, combined with an undersized convolution filter, lead to a slight decrease in performance. \revD{Nevertheless, the confirmation of our hypothesis remains open for future research.}
In the following text, we simplify our notation and use \texttt{RFL} to denote the \texttt{RFL\_Conv} model combined with modalities of set~\ref{row:M032[dcm]S[OApeAprAdT]}.}

\section{Comparison with State-of-the-Art}
\label{subsec:compared-methods}

\begin{table*}[t!]
\centering
\caption{\revB{Comparison of \revDp{the} evaluated label placement methods. The desirable properties are depicted in green, and less desirable properties are marked in orange and red. The computation time requirements are contingent upon the specific use case. For scenarios where labeling needs to be pre-computed with a high level of completeness, such as cartographic maps, technical drawings, or medical atlases, the RFL method is an appropriate choice. Conversely, in interactive applications such as games or viewers, the trade-off of decreased labeling completeness may be acceptable to ensure faster computation times.
 Please note that the \emph{Position model} for the \texttt{PBL-AD} and \texttt{PBL-A} methods is \emph{Fixed/Slider} as these methods first try to position labels on the fixed positions, and only if that is not possible, they try several positions between the fixed ones.} 
}
\label{tab:pros-cons}
\begin{adjustbox}{width=\textwidth}
\begin{tabu}{lllll}
    \toprule
    & \textbf{\texttt{RFL}}  & \textbf{\texttt{RAPL}} & \textbf{\texttt{PBL-AD}}  & \textbf{\texttt{PBL-A}} \\
    \midrule
    \textbf{Paradigm} & Machine Learning & Algorithm & Algorithm & Algorithm  \\
    \textbf{Position model} & Slider & Fixed & Fixed/Slider & Fixed/Slider  \\
    \textbf{Problem definition} & Reward & Value based on rules & Rules & Rules  \\
    \textbf{Label position} & \textcolor{ForestGreen}{Continuous} & \textcolor{BrickRed}{Discrete} & \textcolor{BrickRed}{Discrete} & \textcolor{BrickRed}{Discrete} \\
    \textbf{Heuristic}  & \textcolor{ForestGreen}{Non-greedy} & \textcolor{BrickRed}{Greedy} & \textcolor{BrickRed}{Greedy} & \textcolor{BrickRed}{Greedy}  \\
    \textbf{Computation time} & \textcolor{BrickRed}{Non-interactive} & \textcolor{ForestGreen}{Highly interactive} & \textcolor{Orange}{Interactive (when using collision particles)} & \textcolor{Orange}{Interactive (when using collision particles)} \\
    \textbf{Completeness} & \textcolor{ForestGreen}{High} & \textcolor{BrickRed}{Low} & \textcolor{Orange}{High (achieved by distant labels)}  & \textcolor{BrickRed}{Low}  \\
    \textbf{Resolution-independent} & \textcolor{ForestGreen}{Yes} & \textcolor{BrickRed}{No} & \textcolor{ForestGreen}{Yes} & \textcolor{ForestGreen}{Yes}  \\
    \bottomrule
\end{tabu}
\end{adjustbox}
\end{table*}

\revB{\revC{We employ the completeness metric and benchmark dataset, as outlined in \autoref{subsec:completeness-metric} and \autoref{subsec:benchmark-dataset}, to compare our Reinforced Labels (\texttt{RFL}) with several published methods, beginning} with the implementation of Particle-Based Labeling (\texttt{PBL}) proposed by Luboschik~\etal~\cite{Luboschik2008}. The latter method attempts to position each label sequentially, first with the fixed 4- and 8-position model, then using the slider model, and finally with a spiral-based distant model.
Furthermore, the method is greedy, and such cannot change the position of a label after it has been placed. Moreover, labels that the method cannot position without a conflict are removed from the calculated label layout. Therefore, the approach may produce \revDg{an} incomplete label layout.
We separate \texttt{PBL} into two variants. The first, denoted as \texttt{PBL-A}, involves only \revBa{the fixed 4-, 8-position, and slider} models resulting in \revB{\textbf{adjacent}} label placement. 
The other variant, referred to as \texttt{PBL-AD}, applies \revBa{the spiral-based distant model in addition to the fixed 4-, 8-position, and slider} models, resulting in a combination of both \textbf{adjacent and distant} label placement (\ie a label can be placed farther away from its anchor, and a leader line maintains the correspondence). 
\revC{Please be aware that \texttt{PBL-AD} cannot be directly compared with adjacent-only methods, such as \texttt{RFL}, given the fact that the distant labels offer a greater degree of freedom. As a result, we expect \texttt{PBL-AD} to achieve a higher completeness score than adjacent-only methods. Nevertheless, we include the \texttt{PBL-AD} in the evaluation to compare the \texttt{RFL} with a method that uses distant labels.}
}

\revB{%
Additionally, we compare \texttt{RFL} to the Rapid Labels (\texttt{RAPL})~\cite{Pavlovec2022}, a GPU-accelerated greedy and adjacent-only method that leverages the %
8-position model. 
Similarly to \texttt{PBL}, \texttt{RAPL} may produce \revDg{an} incomplete layout as labels that \revDr{the method cannot place without a conflict}\revDp{cannot be placed without a conflict by the method} are removed from the calculated layout. \autoref{tab:pros-cons} presents a comparison of the evaluated label placement methods.}

\revA{Finally, we introduce an untrained version of \texttt{RFL} with randomly initialized weights (abbreviated as \texttt{RFL-random}) to validate that the \texttt{RFL} learns a reasonable policy.}
Comparison with \revDp{the} other existing methods\revD{, as discussed in \autoref{sec:related-work},} is infeasible due to \revDp{the} unpublished or proprietary implementations, which we were unable to get or re-implement.

\subsection{Quantitative Results}
\label{subsec:quantitative-results}

\begin{figure*}[t!]
    \subfigure[] {
        \label{fig:compact-completeness}
        \includegraphics[width=0.3197\textwidth]{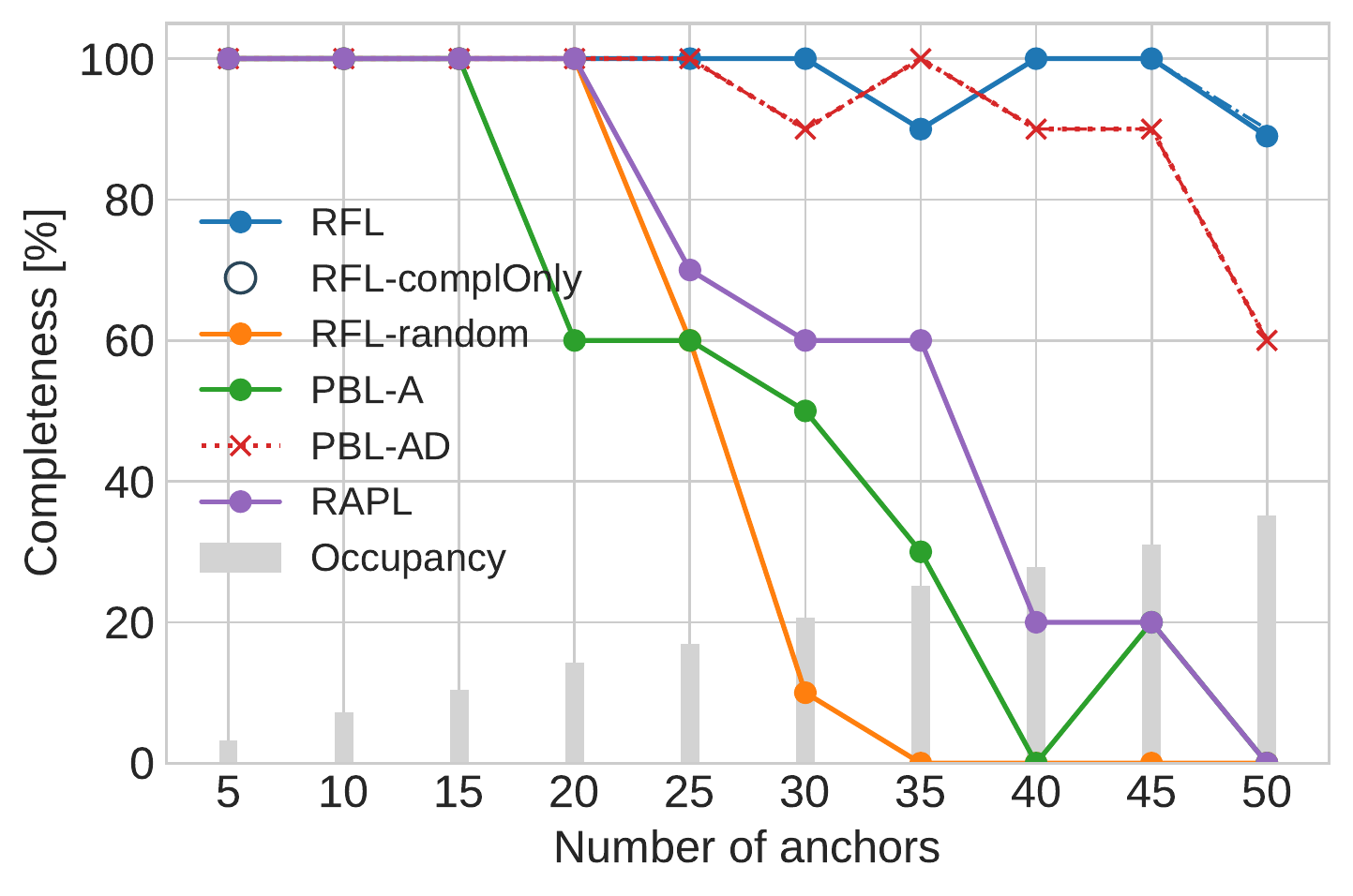}
    }
    \hfill
    \subfigure[] {
        \label{fig:compact-time}
        \includegraphics[width=0.3197\textwidth]{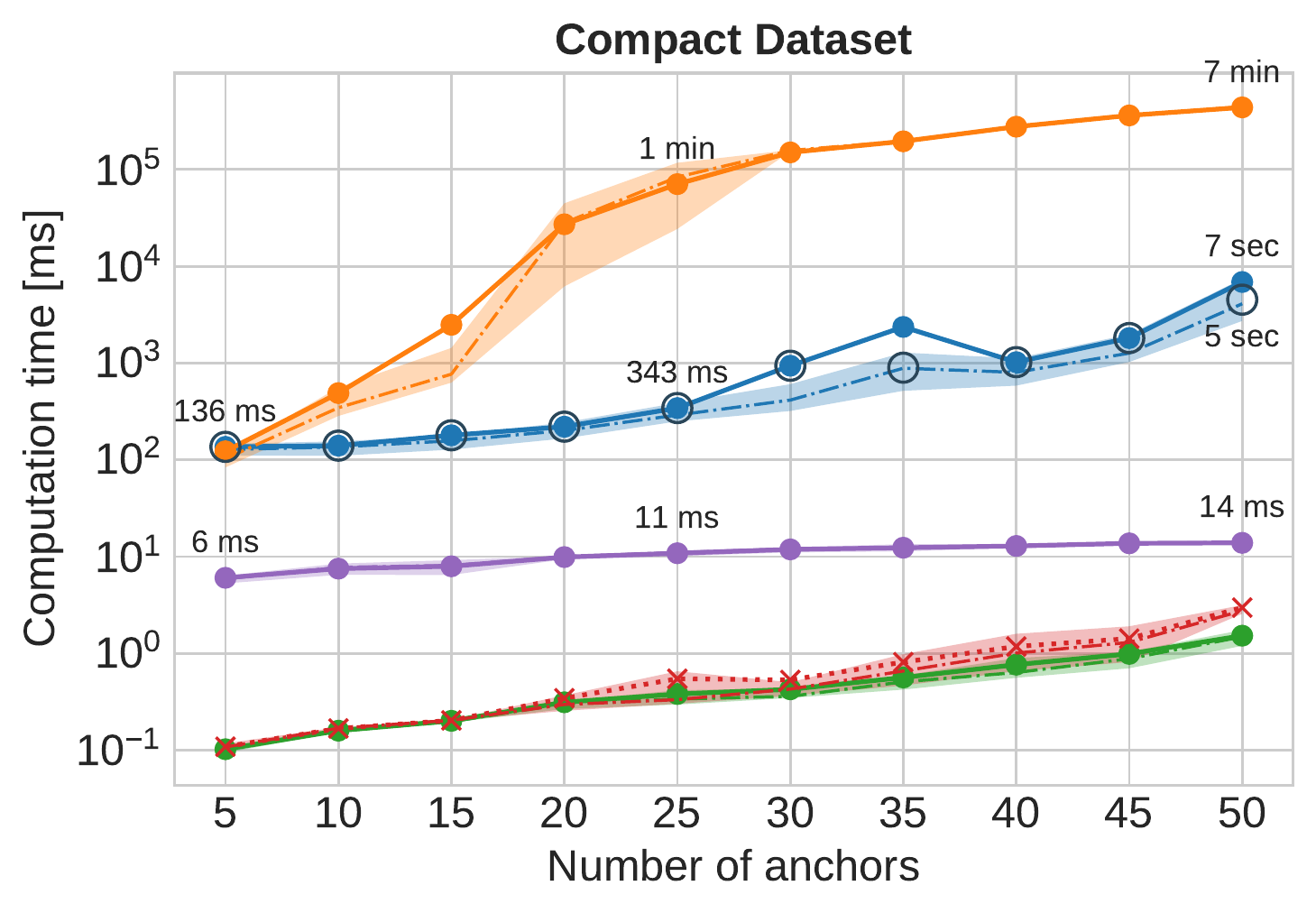}    
    }
    \hfill
    \subfigure[] {
        \label{fig:compact-episode-length}
        \includegraphics[width=0.3197\textwidth]{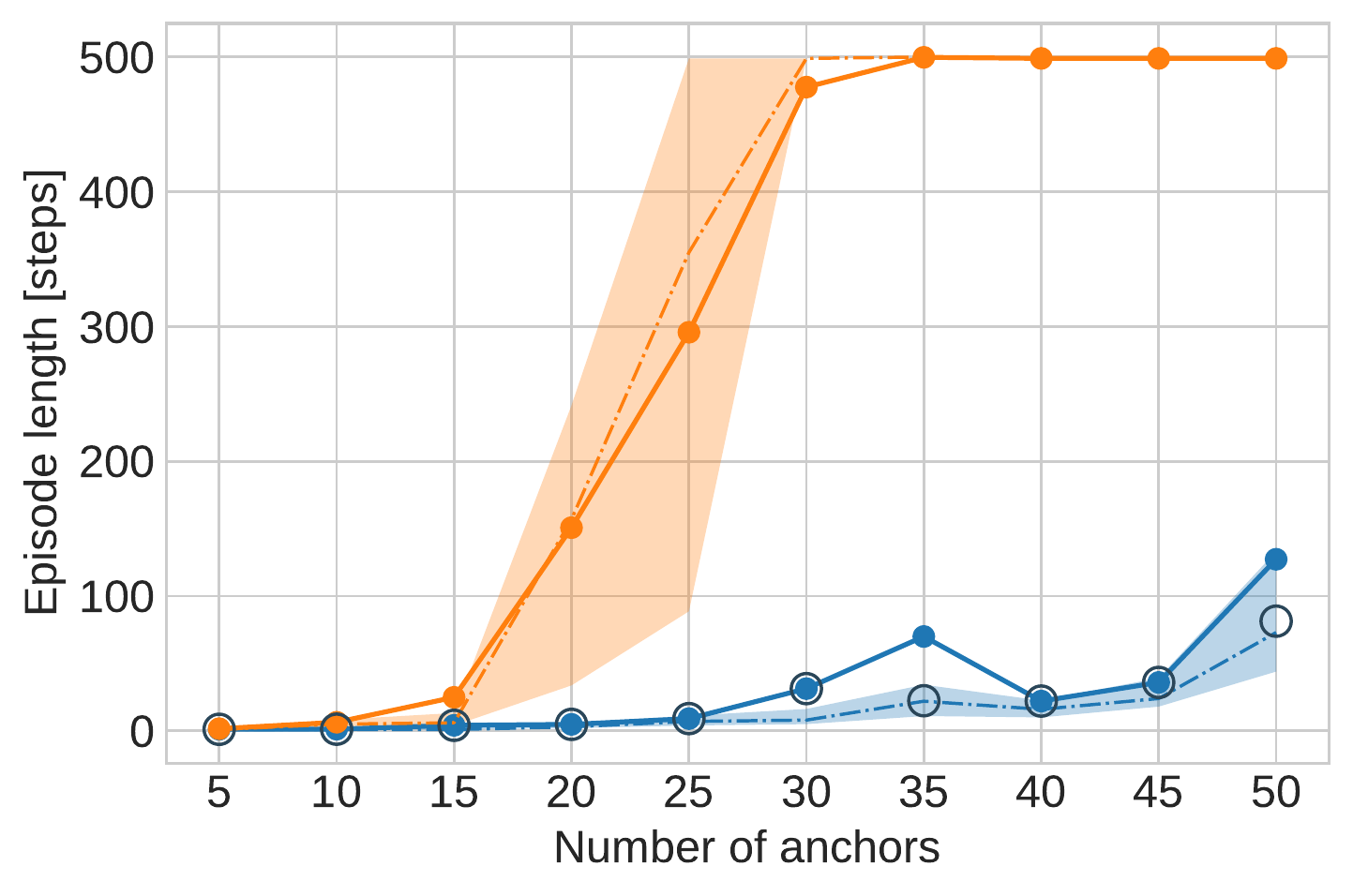}
    }
    \subfigure[] {
        \label{fig:volume-completeness}
        \includegraphics[width=0.3197\textwidth]{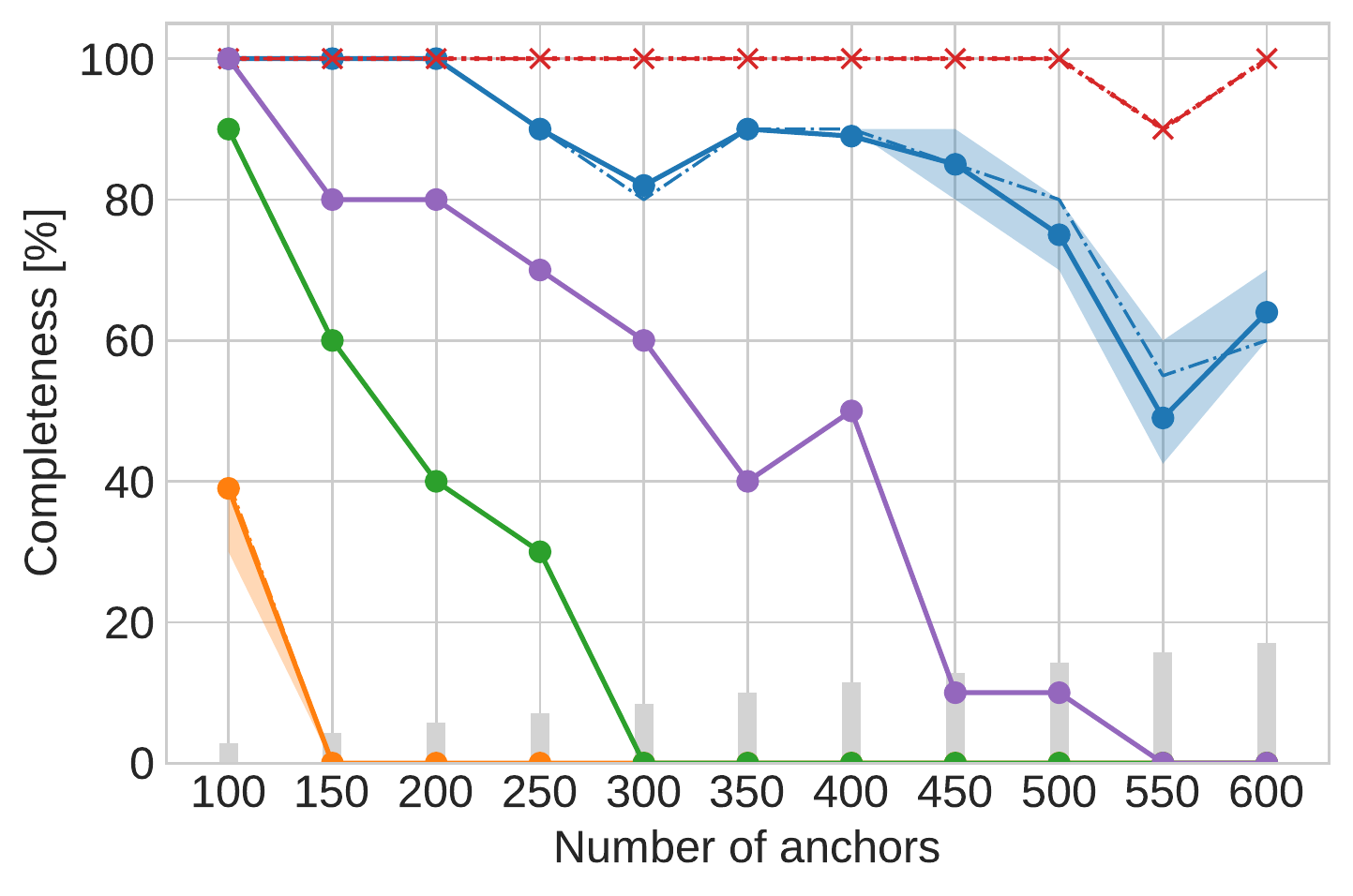}
    }
    \hfill
    \subfigure[] {
        \label{fig:volume-time}
        \includegraphics[width=0.3197\textwidth]{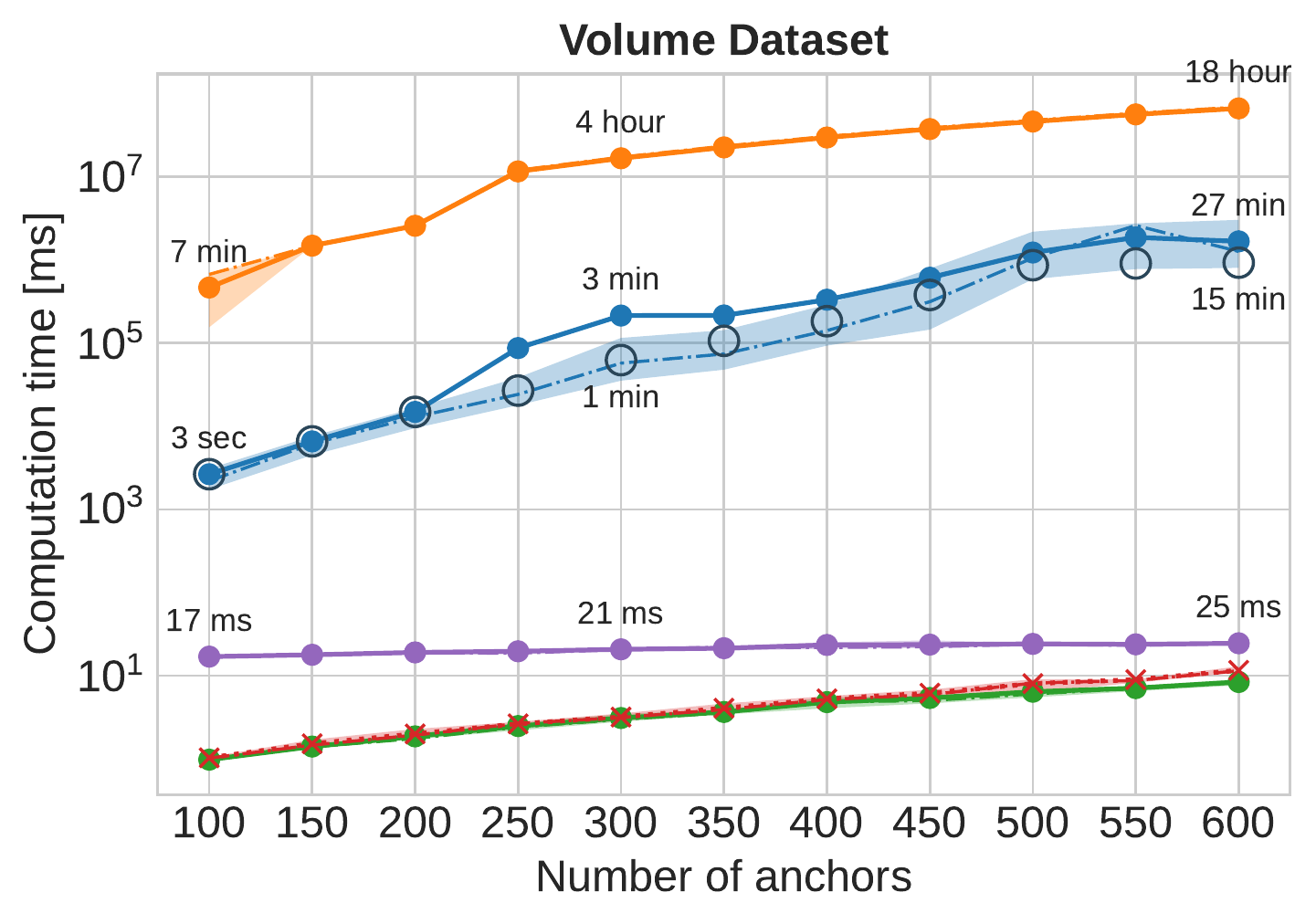}    
    }
    \hfill
    \subfigure[] {
        \label{fig:volume-episode-length}
        \includegraphics[width=0.3197\textwidth]{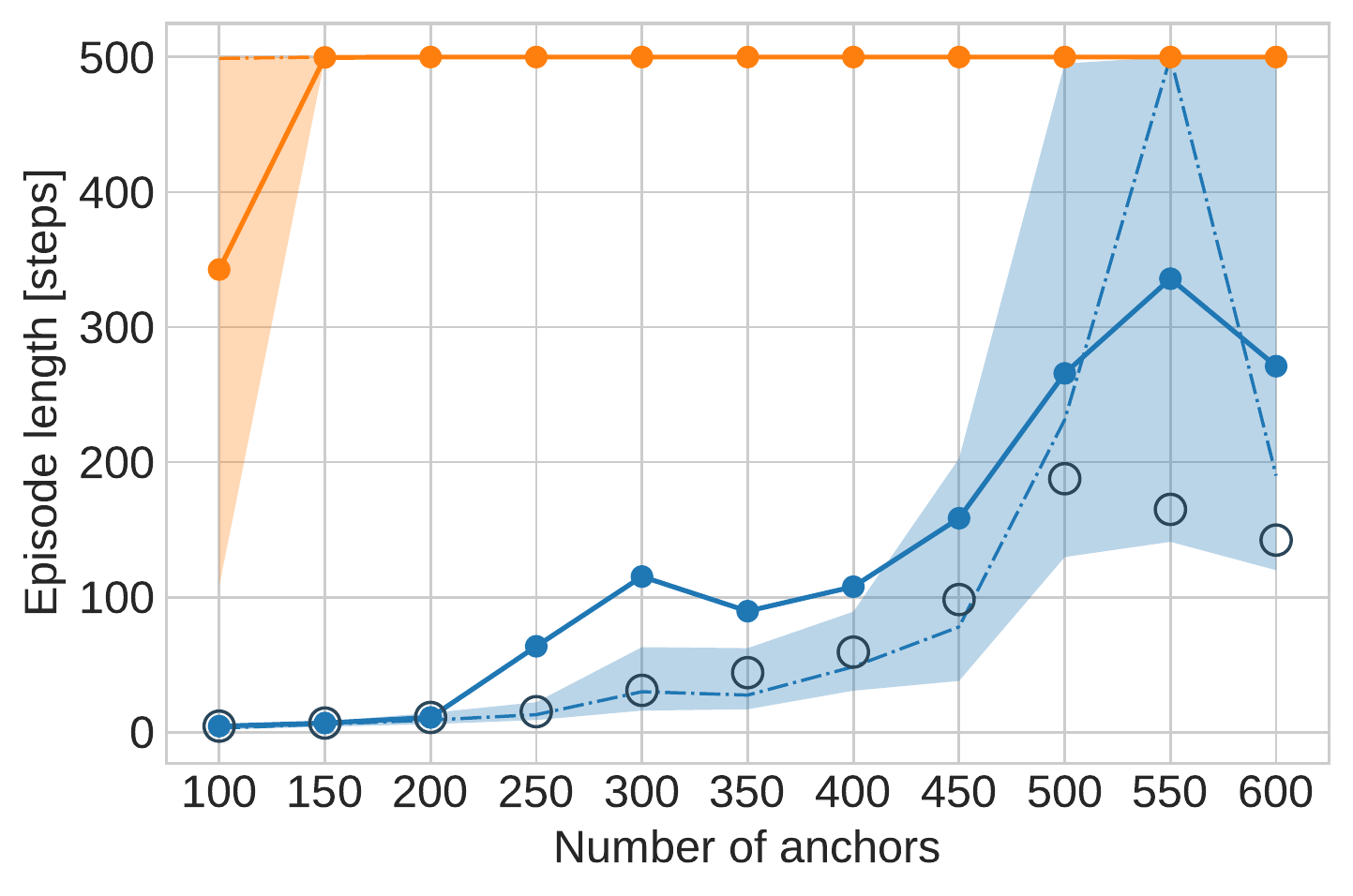}
    }
    \caption{Comparison of \revDp{the} examined methods evaluated on the benchmark dataset. Charts \subref{fig:compact-completeness} and \subref{fig:volume-completeness} illustrate the completeness on compact and volume datasets. Charts \subref{fig:compact-time} and \subref{fig:volume-time} show computation time. Charts \subref{fig:compact-episode-length} and \subref{fig:volume-episode-length} depict the length of the episode. \revB{We recall that \texttt{PBL-AD} combining both \textbf{adjacent and distant} models cannot be directly compared with \textbf{adjacent}-only methods such as \texttt{RFL}, \texttt{PBL-A}, or \texttt{RAPL}.The former \texttt{PBL-AD} provides a greater degree of freedom by placing a label further away from its anchor while the additional leader line maintains the correspondence.} We acquired the values for \texttt{RFL} and \texttt{RFL-random} by executing these methods ten times over a set of instances with the same number of anchors. The solid line represents the mean, the shaded area depicts quartiles Q1 and Q2, and the dash-dot line describes the median. \revB{The empty circle symbol represents the mean value of complete-only instances found by \texttt{RFL} (incomplete instances reaching the fixed horizon are filtered out). In simpler terms, the value conveyed by the symbol provides an answer to the question: "If and only if the \texttt{RFL} can solve the given instances, how long does it take on average?"}}
    \label{fig:comparison}
\end{figure*}

We trained the \texttt{RFL} policy on randomly generated instances previously described in \autoref{subsec:environment} with at the most two anchors, therefore with up \revA{to} two agents within the environment, for \revA{an hour} on a computation node equipped with 2x AMD EPYC\textsuperscript{\texttrademark} 7H12, 64-core, 2.6 GHz CPUs without a GPU accelerator. To optimize the parameters of the neural network, we utilize PPO implementation from the RLLib framework~\cite{liang2018rllib}. Specifically, we used 119 cores for rollout workers to collect agents' experiences (\ie observations and rewards) and a single core for the trainer worker responsible for updating the parameters of the proposed network. A detailed description of our training setup is available in the supplementary material. We executed the following evaluations on Intel$\textsuperscript{\tiny\textregistered}$ Core i7-9700K 8-core, 3.60GHz CPU, and NVIDIA GeForce GTX 1660~Ti.
To capture the stochastic nature of \texttt{RFL} (\revDr{recall}\revDp{see} the definition of \autoref{eq:agent-policy}), we evaluated the \texttt{RFL} and \texttt{RFL-random} ten times over the benchmark dataset. The other compared methods, \texttt{PBL-A}, \texttt{PBL-AD}, and \texttt{RAPL}, are deterministic, and as such, we evaluated each only once. Therefore, we provide quartiles Q1 and Q2, mean and median statistics for the completeness metric of \texttt{RFL}. Furthermore, we fixate the episode horizon of \texttt{RFL} and \texttt{RFL-random} in the evaluation phase at $T=500$ steps (recall we set the $T=100$ for the training).

\autoref{fig:compact-completeness} and \autoref{fig:volume-completeness} show the completeness of \revDp{the} compared methods on the compact and volume datasets. Both charts reveal a similar trend. \revB{As the number of anchors rises, as \revDr{well as}\revDp{does} the environment \textit{occupancy} (\ie a ratio of the total area of the labels to the overall area of the drawing), the completeness of \texttt{PBL-A}, \texttt{RAPL}, and \texttt{RFL-random} decreases rapidly.}
The reason behind the difference in completeness between 50 anchors in the compact and 100 anchors in the volume dataset is the varying occupancy. It is easier to find complete labeling within a larger space of volume dataset.
Besides \texttt{PBL-AD}, which has the advantage of distant labels,
\revDr{the best performs \texttt{RFL}, achieving 89\% and  64\% completeness on average for 50 and 600 anchors.}\revD{the \texttt{RFL} achieves the highest average completeness of 89\% and 64\% for 50 and 600 anchors, respectively.} 
The random policy of an untrained agent \texttt{RFL-random} corresponding with a chance performs the worst. This fact confirms that the \texttt{RFL} learns a meaningful policy, and its performance is not the outcome of the randomized search.
The second worst method \revD{in terms of completeness} is \texttt{PBL-A}, followed by \texttt{RAPL}. \revBa{Starting from 300 anchors, \texttt{PBL-A} fails to place all the labels for any instance, resulting in 0\% completeness. Similarly, at 50 and 550 anchors, \texttt{PBL-A}, \texttt{RAPL}, and \texttt{RFL-random} all achieve 0\% completeness.}

We also compared \texttt{RFL} with \texttt{PBL-AD}, which combines \textbf{adjacent and distant} labels and, therefore, has a higher degree of freedom than the strict slider model of our method. 
\revD{However, it is essential to recognize that the inherent flexibility of \texttt{PBL-AD}, stemming from its ability to position distant labels away from the anchors and \revDp{the} use of leader lines for connection, makes it directly incomparable with adjacent-only methods such as \texttt{RFL}.}
\revD{Moreover, the leader lines that grant \texttt{PBL-AD} its flexibility can also intersect with other labels, anchors, or leader lines, creating potential conflicts. Such conflicts not only diminish the labels' readability but also complicate the association between labels and anchors. Imhof~\cite{Imhof1975} underscores the importance of both label readability and straightforward label-to-anchor association in label layout. Hence, while \texttt{PBL-AD} might seem advantageous regarding placement flexibility, it should be approached cautiously, especially when label layout quality is essential.}
To our surprise, the \texttt{PBL-AD} performs slightly worse on the compact dataset than \texttt{RFL} and surpasses \texttt{RFL} only at the point of 35 anchors. We further investigated this case and found that the dip in \texttt{RFL} performance is caused by the factual inexistence of a complete conflict-free layout. Therefore, the proposed method outperforms both hand-crafted algorithms \texttt{RAPL} and \texttt{PBL-A}/\texttt{-AD} on the compact dataset. On the volume dataset, the \texttt{PBL-AD} shows stable completeness of 100\% through the dataset, except at 550 anchors, where the completeness dips slightly to 90\%.
\revC{We attribute the superior performance of \texttt{PBL-AD} on the volume dataset to the fact that anchors are not spread as evenly across the entire space as in the compact dataset. As a result, \texttt{PBL-AD} can utilize more distant labels for anchors in dense clusters and position them in less dense areas.}

\autoref{fig:compact-time} and \autoref{fig:volume-time} depict the dependence of the computation time on the number of anchors. We group all instances with a given number of anchors and compute \revDp{the} aggregated statistics. Therefore, we report quartiles Q1 and Q2, mean and median statistics. The \texttt{PBL-A} and \texttt{PBL-AD} are the fastest methods over the entire benchmark dataset. The \texttt{RAPL} follows with a difference of an order of magnitude that steadily decreases towards 600 anchors. This fact goes along with the authors' statement that the performance gain comes with a more significant number of anchors due to the computation of \revDp{the} Summed Area Table~\cite{Pavlovec2022}.
\revDr{Second-to-last performs the \texttt{RFL}}\revD{The second slowest method is \texttt{RFL}}.
\revA{%
We attribute this to the RLLib's internal inefficiencies (\ie policies among agents within an environment cannot be evaluated in parallel) and the utilized single-thread ray casting implementation from the Box2D framework\footnote{The framework Box2D is available at \url{https://box2d.org}.}.
In fact, the observation computation and collection take on average 46\% (29~ms) of the computation time per step, of which 2/3 makes up the ray-casting operation. The inference of the proposed architecture carries only about 7\% (4 ms) of the time. The final 47\% (30~ms) is dissolved in preprocessing observations and actions within the RLLib framework.
}%
\revDr{The \texttt{RFL-random} performs the worst}\revD{The slowest method overall was \texttt{RFL-random}}, hitting the upper bound of 500 steps. Again, this fact confirms that \texttt{RFL} learns a reasonable policy, and its performance is not an outcome of the randomized search.

\autoref{fig:compact-episode-length} and \autoref{fig:volume-episode-length} illustrate the dependence of \texttt{RFL} and \texttt{RFL-random} \revDp{on} episode length and the number of anchors. The results significantly distinguish the policy of trained and untrained agents corresponding with a chance that a random set of actions reach the complete conflict-free labeling. The difference is mainly visible at 30 and 150 anchors -- the policy of \texttt{RFL-random} skyrockets to the horizon of 500 steps while \texttt{RFL} tops 32 and 7 steps at the same point, respectively.

\begin{table*}[!t]
\caption{
\revA{%
Visual comparison of \revDp{the} examined methods for selected instances from the proposed dataset. In addition, we provide real-world instances of IATA airport codes with 250 anchors and CITY names with 150 anchors based on data obtained from Open Street Maps. For the latter two, we cropped the result to focus on differences among methods. Original results can be found in the supplementary material.
The green dot represents the anchor (\ie point feature). The gray rectangle symbolizes the body of the label itself. The red dot describes an anchor that was not labeled by the given method. The red rectangle illustrates the dimensions of the missing label. We stress that we intentionally added all the missing labels to the visualization for illustrative purposes only, and their origins are not the outcome of the method itself.}
}
\label{tab:visual-comparison}
\scriptsize
\centering
\begin{tabu}{c  c  c  c  c }
     \toprule
      \rotatebox{90}{\hspace{-0.8cm}Anchors} & \texttt{RFL} & \texttt{RAPL} & \texttt{PBL-A} & \texttt{PBL-AD} \\
      \\ 
      \rotatebox{90}{\hspace{1.1cm}45}
      &
      \includegraphics[width=0.22\textwidth, frame]{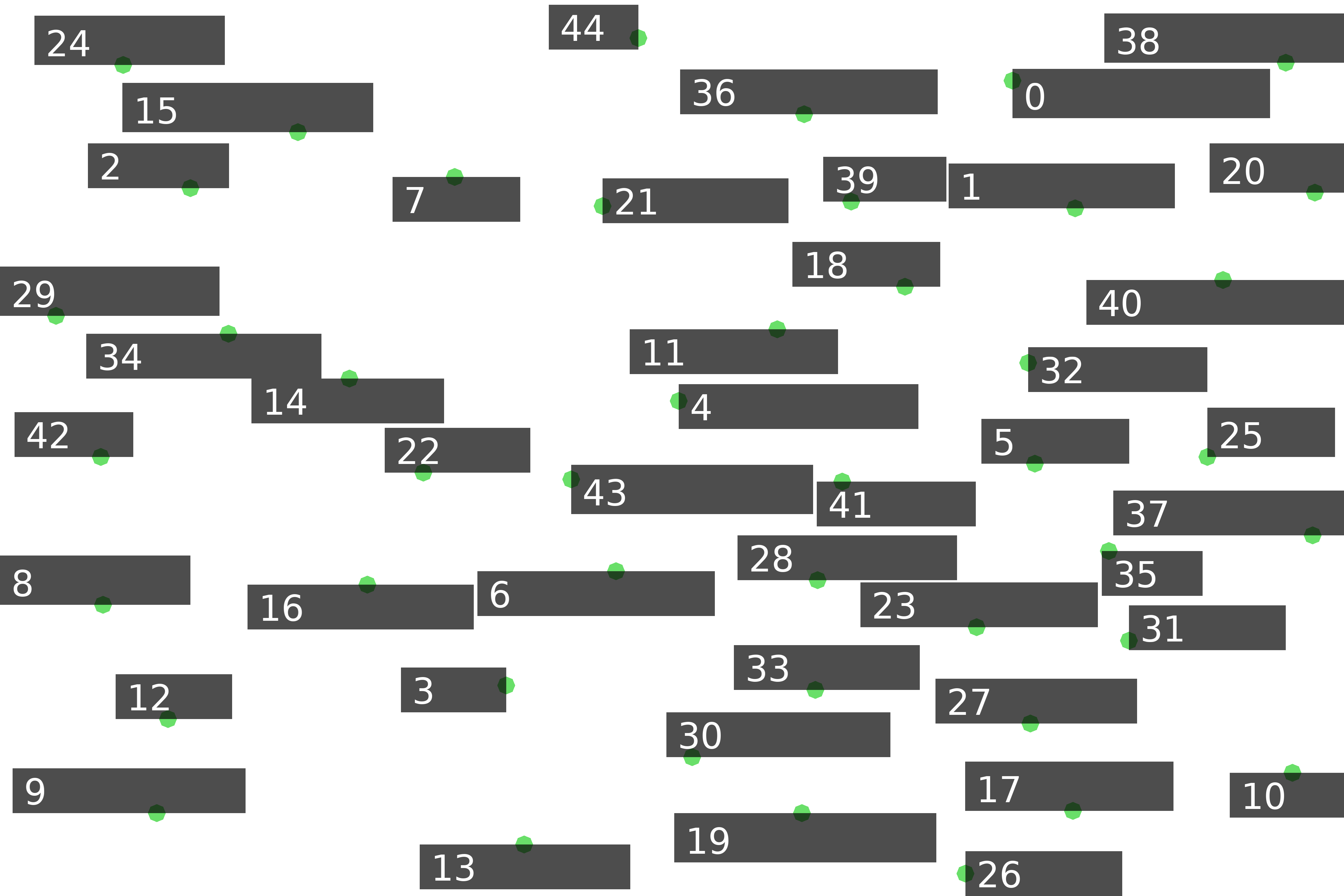}
      & 
      \includegraphics[width=0.22\textwidth, frame]{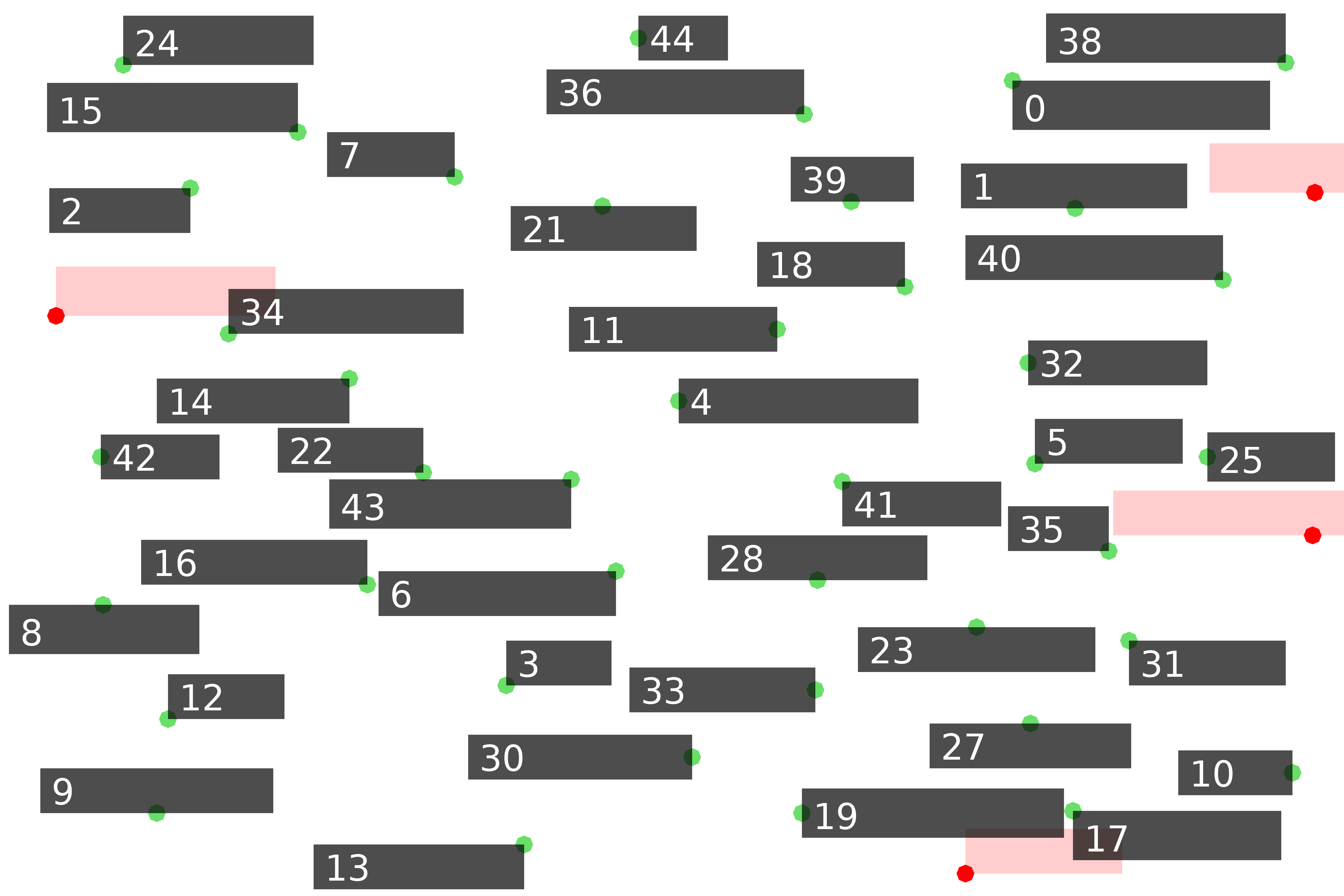}
      & 
      \includegraphics[width=0.22\textwidth, frame]{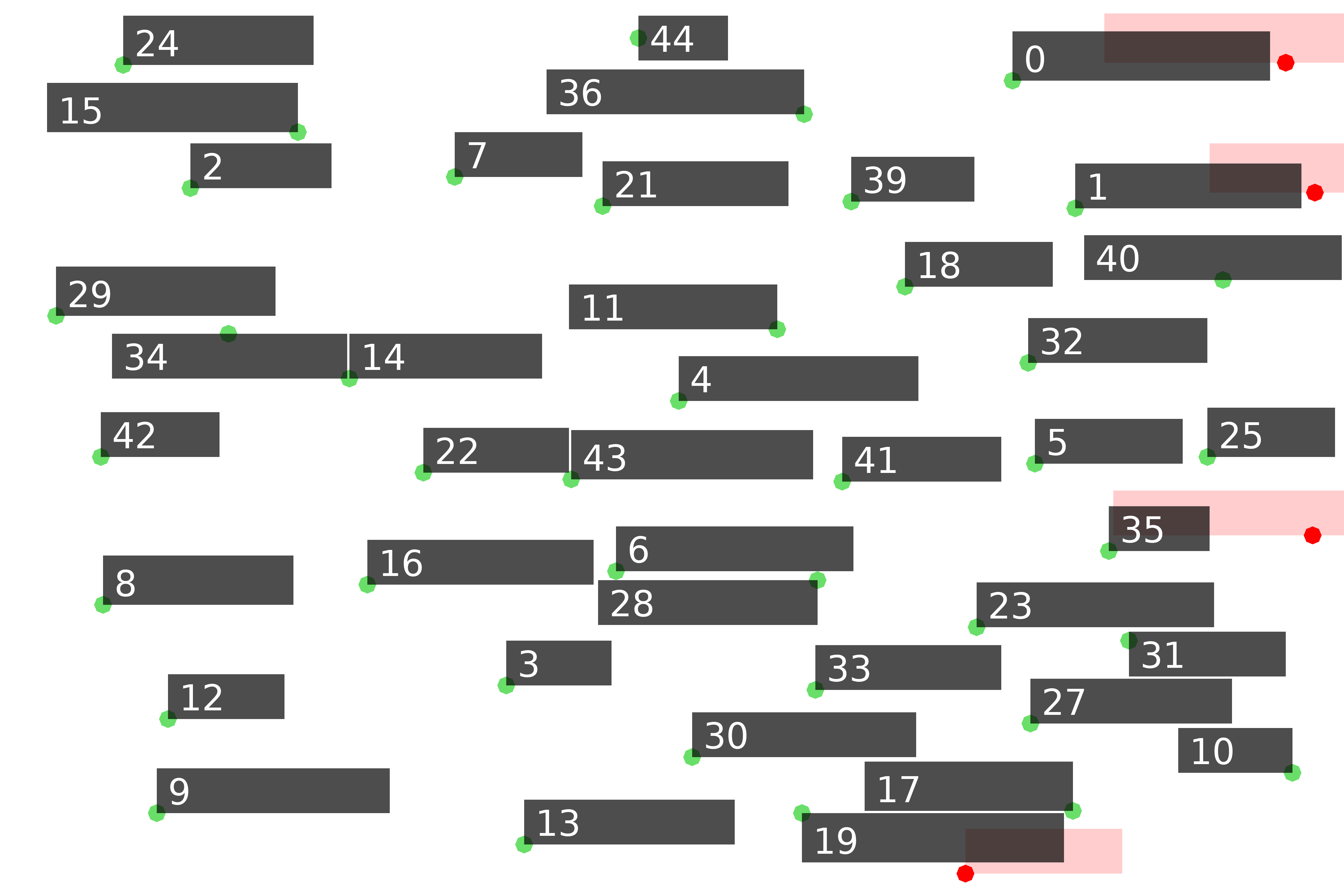}
      & 
      \includegraphics[width=0.22\textwidth, frame]{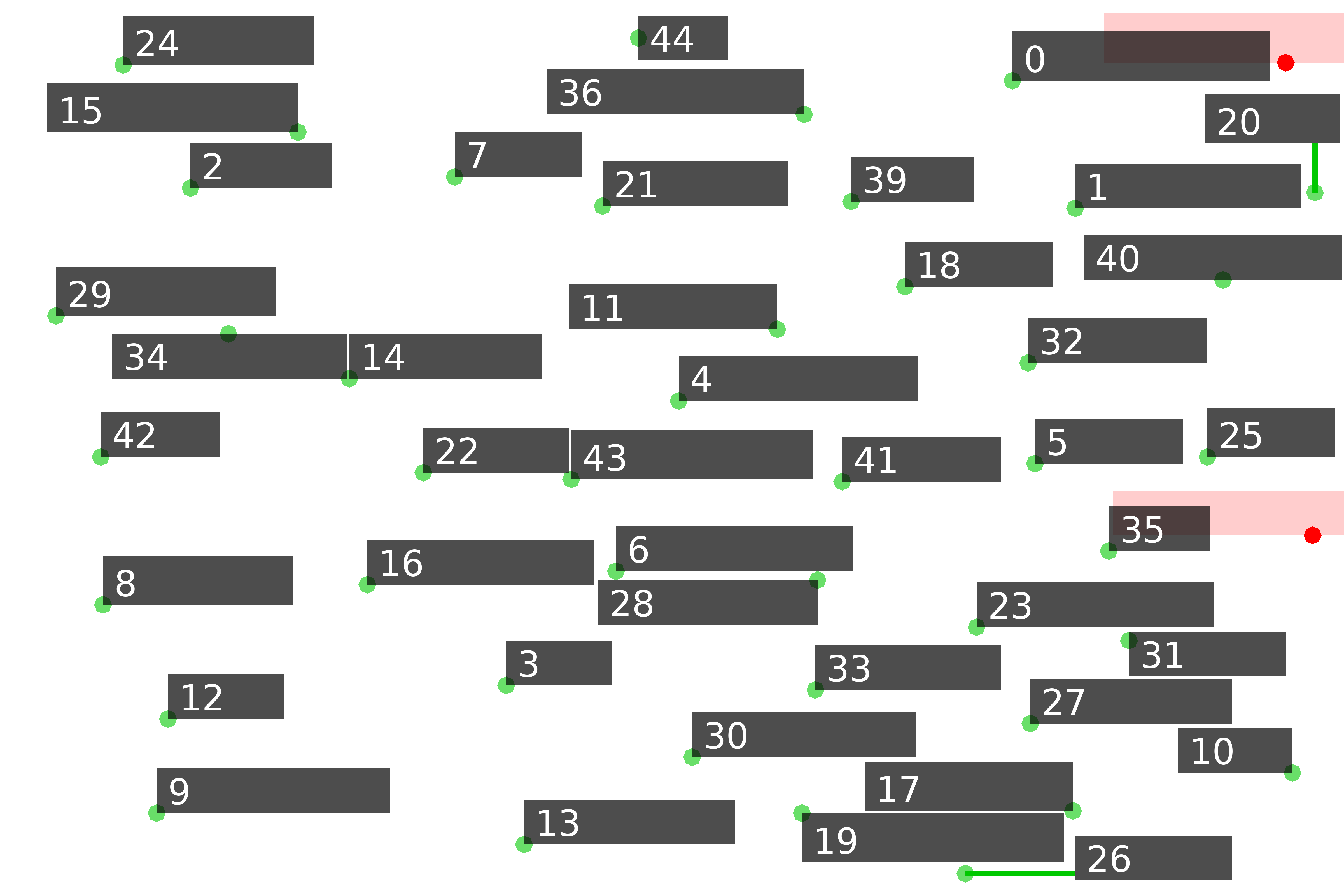}
      \\
      \rotatebox{90}{\hspace{1.1cm}150 CITY}
      &
      \includegraphics[width=0.22\textwidth, frame]{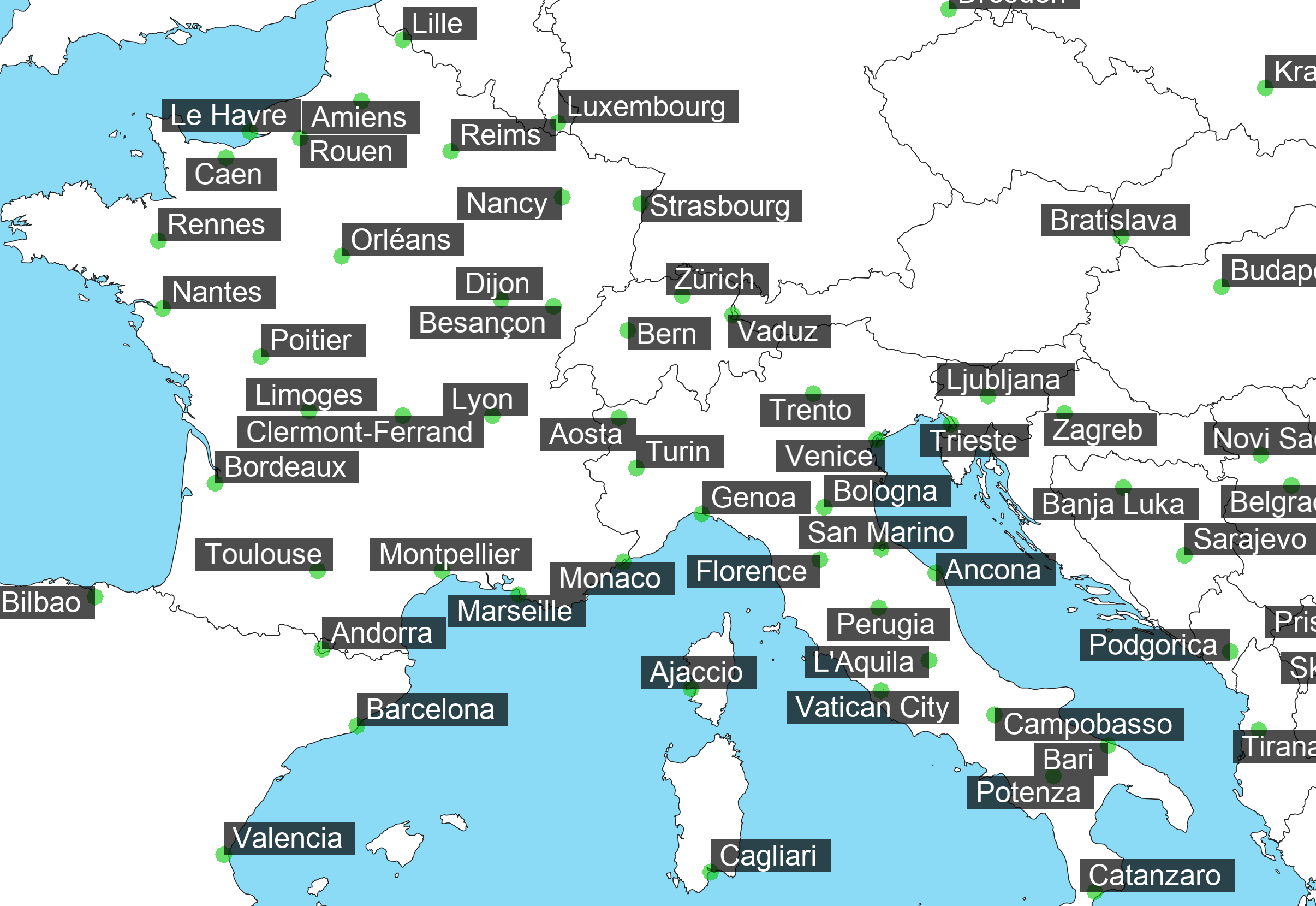}
      & 
      \includegraphics[width=0.22\textwidth, frame]{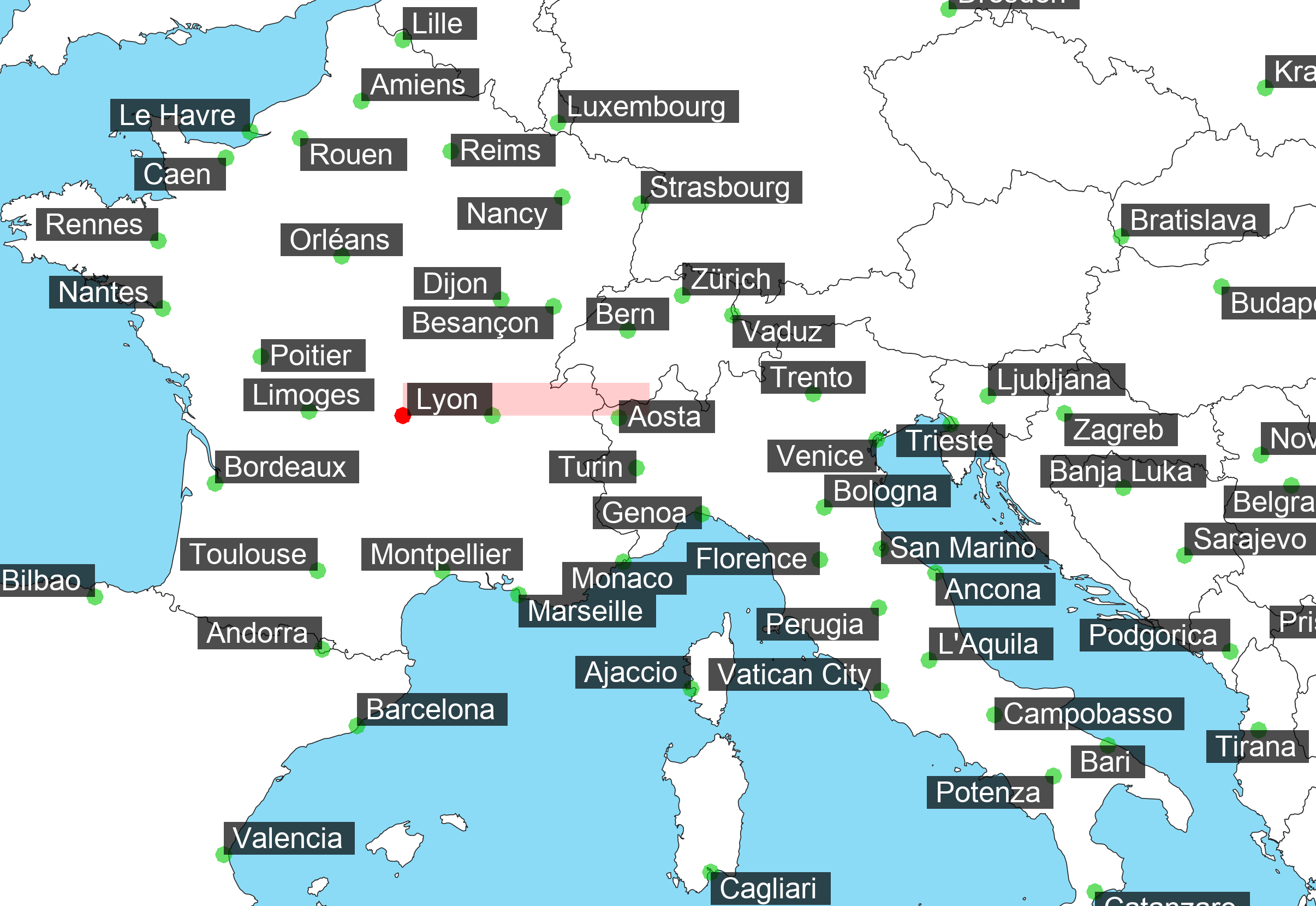}
      & 
      \includegraphics[width=0.22\textwidth, frame]{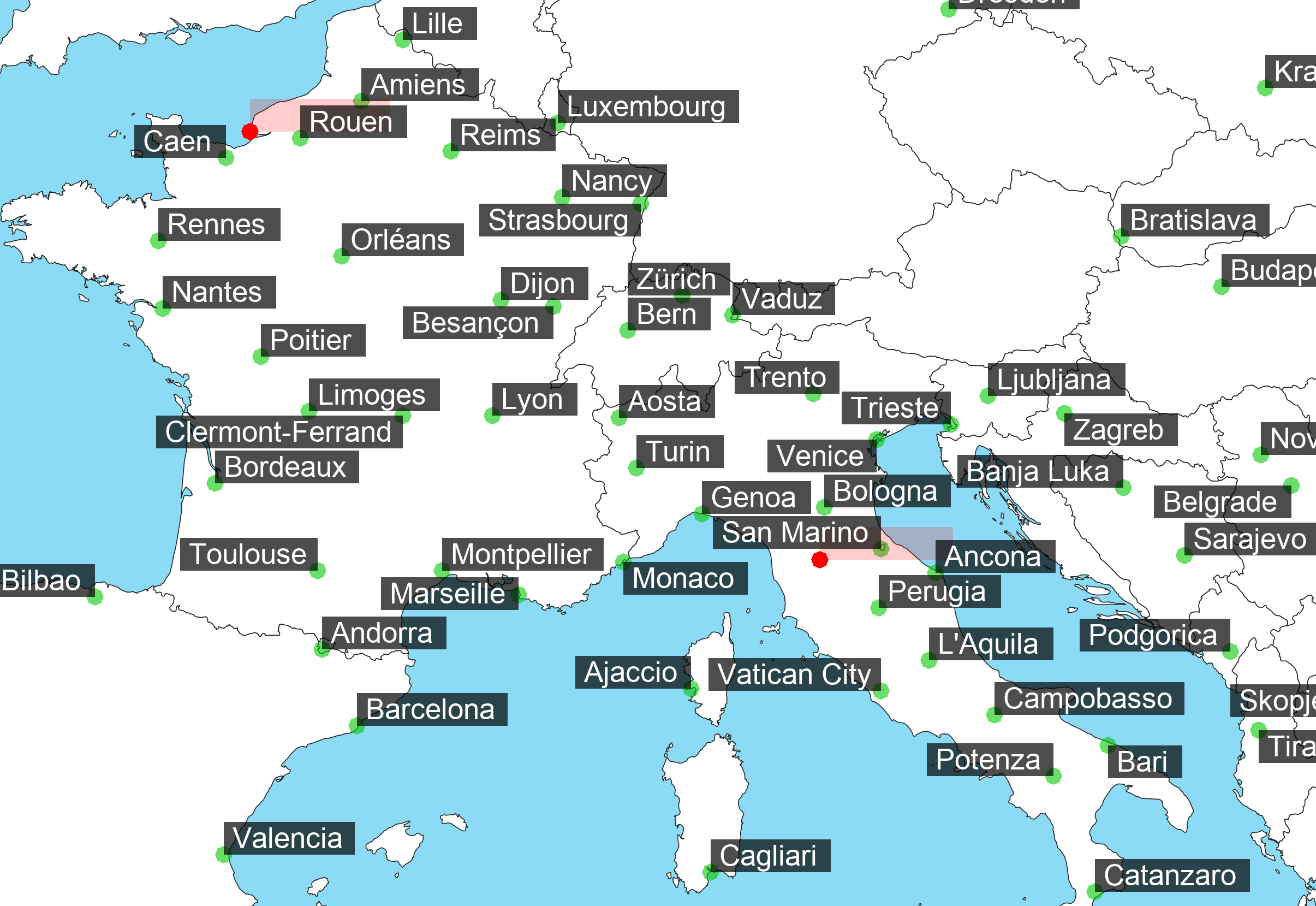}
      & 
      \includegraphics[width=0.22\textwidth, frame]{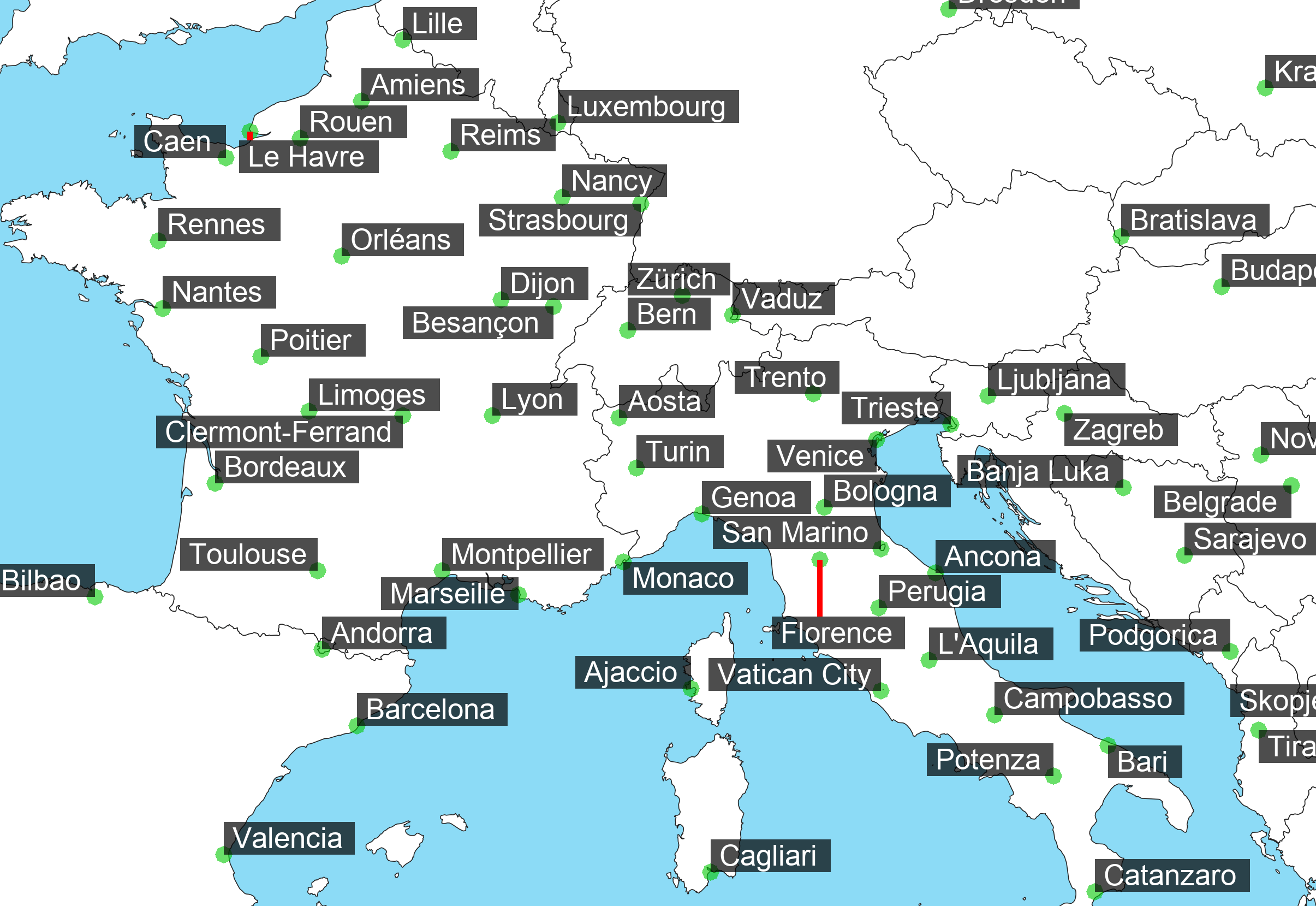}
      \\
      \rotatebox{90}{\hspace{1.1cm}250 IATA}
      &
      \includegraphics[width=0.22\textwidth, frame]{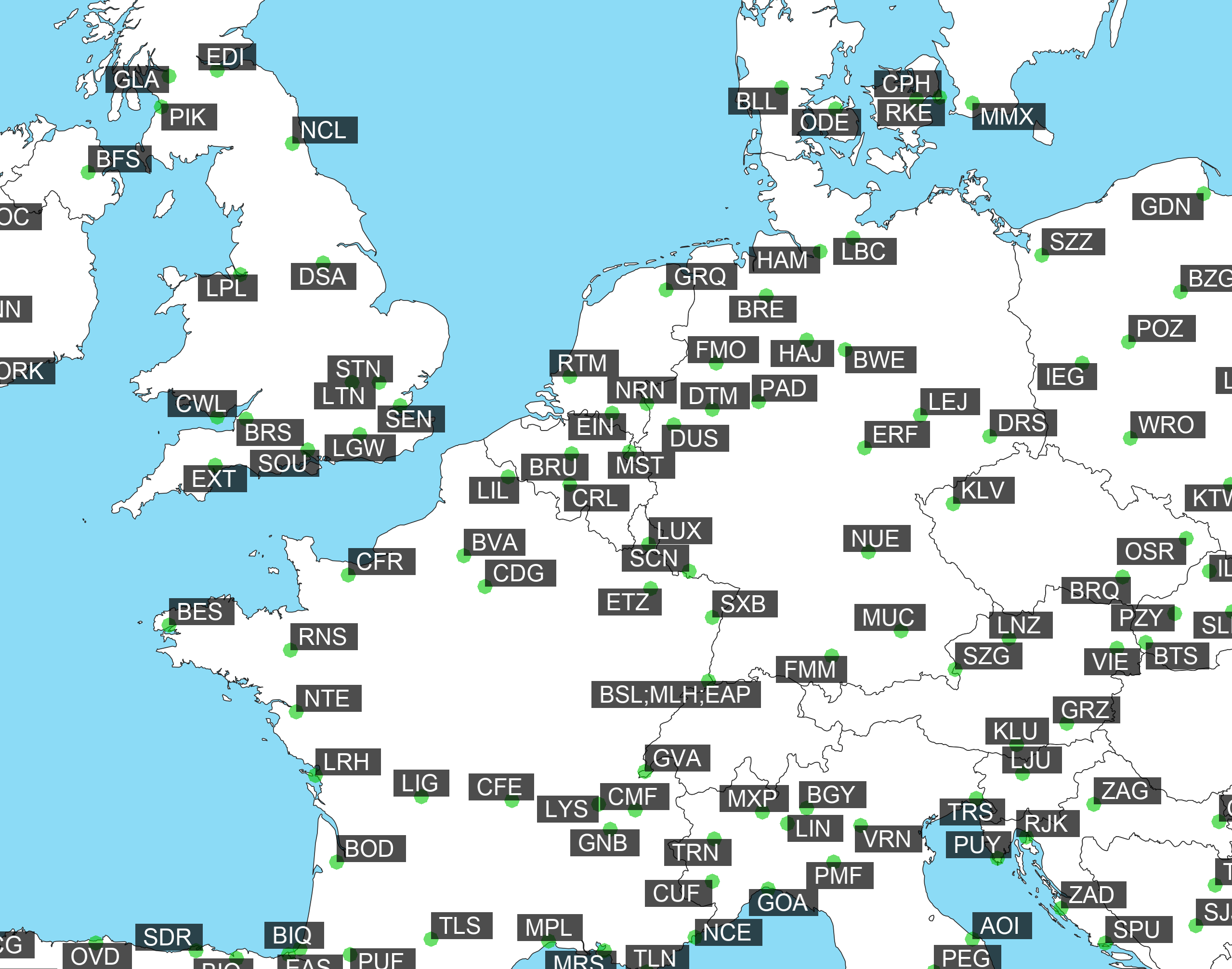}
      & 
      \includegraphics[width=0.22\textwidth, frame]{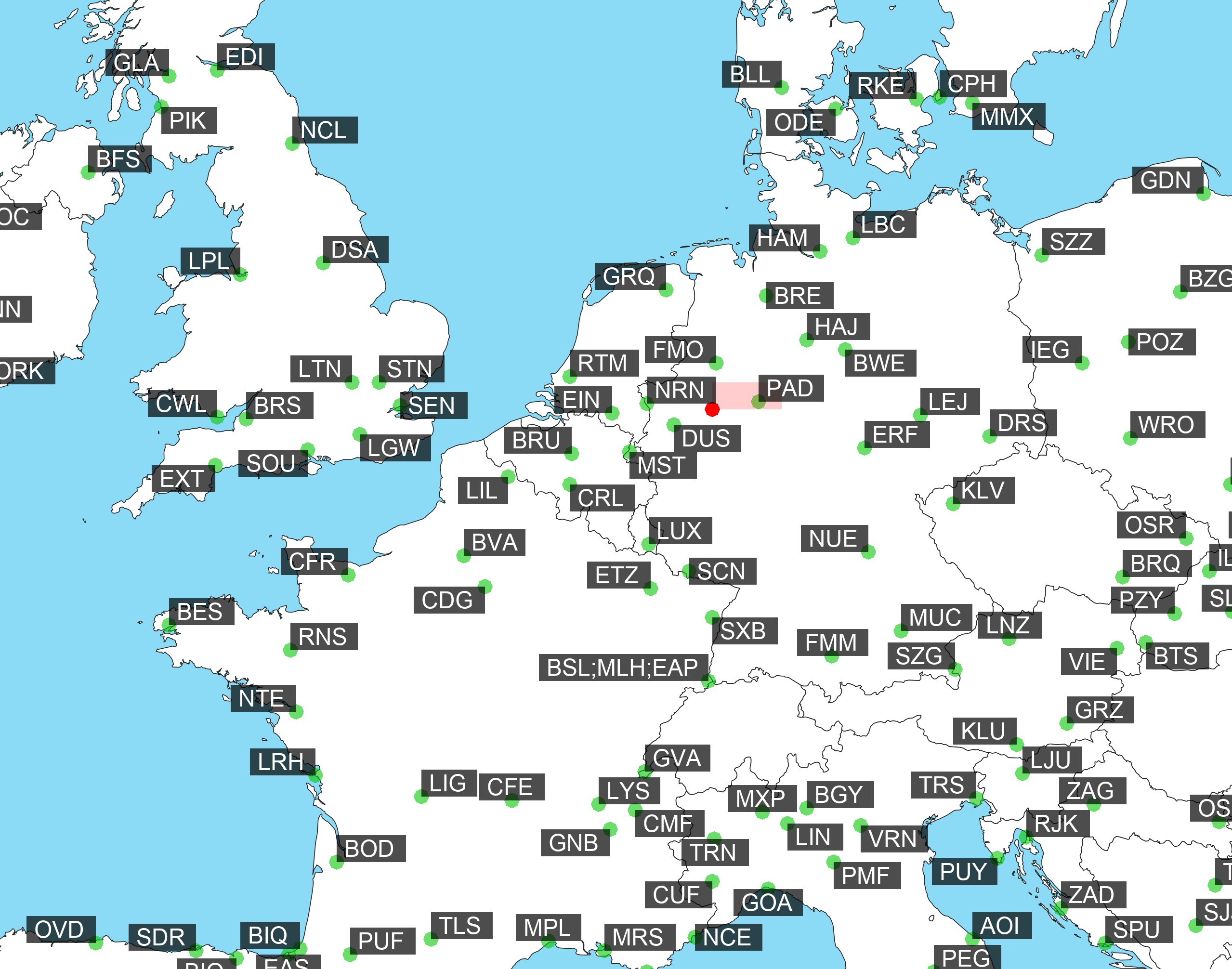}
      & 
      \includegraphics[width=0.22\textwidth, frame]{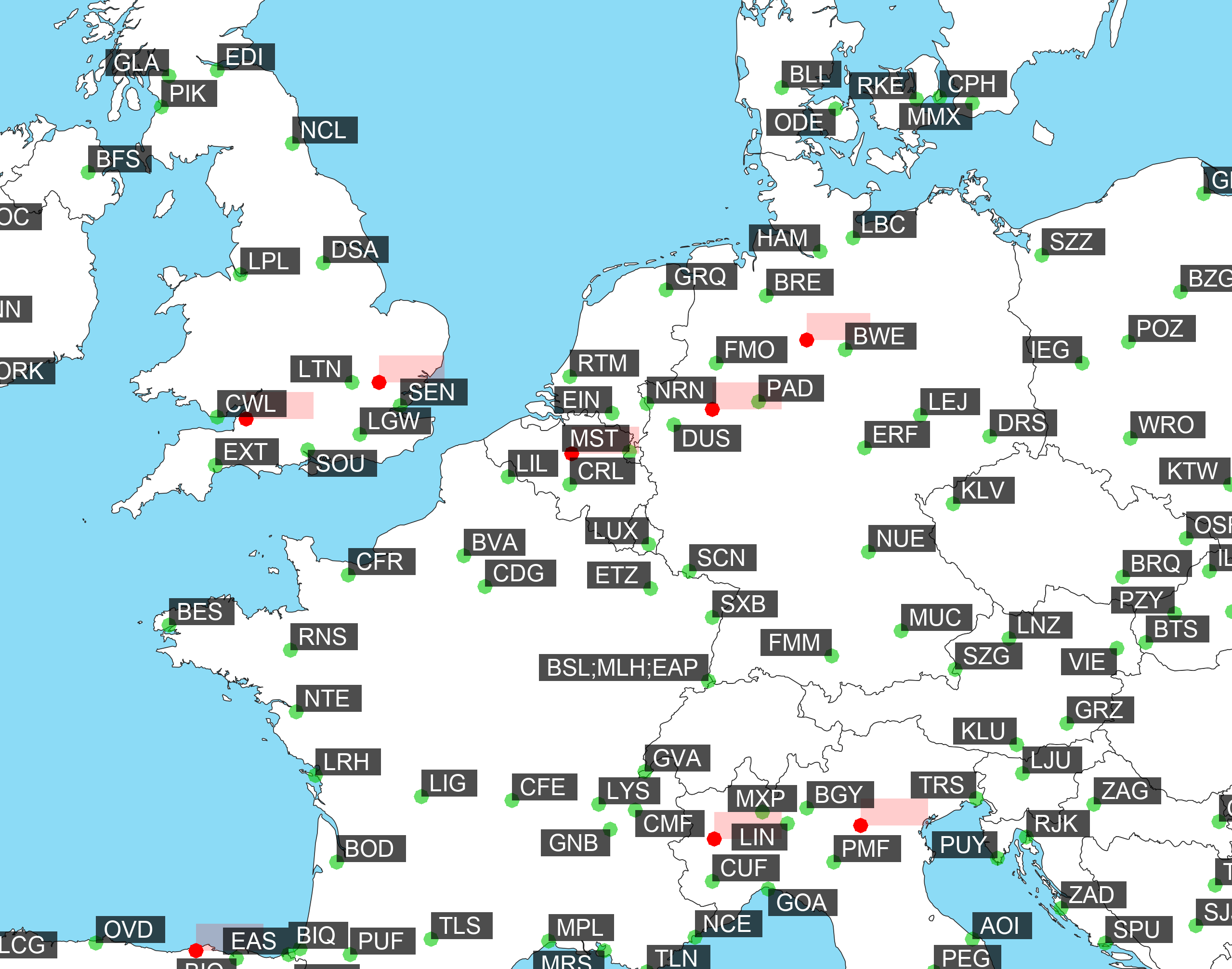}
      & 
      \includegraphics[width=0.22\textwidth, frame]{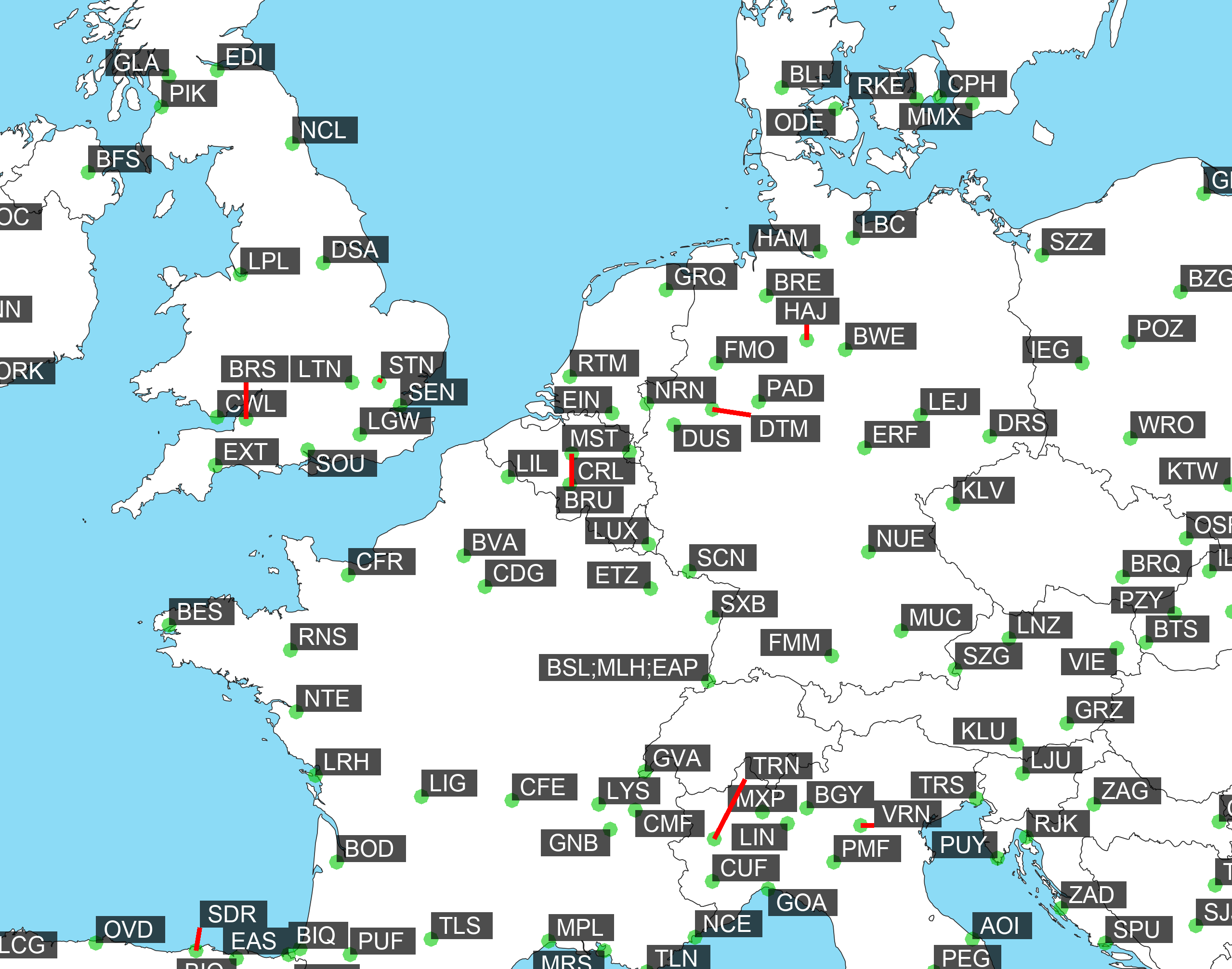}
      \\
\end{tabu}
\end{table*}

\subsection{User Study}
\label{subsec:user-study}
\revB{Along with the quantitative evaluation, we conducted a user study to determine the preferred methods among users. We assessed the \texttt{RFL} method, as well as the state-of-the-art methods \texttt{RAPL}, \texttt{PBL-A}, and \texttt{PBL-AD}, which were also evaluated in the quantitative analysis.}
\revB{To carry out the user study, we utilized instances from the compact dataset containing 30, 40, and 50\revDr{ randomly placed} anchors and labeled each instance with all the evaluated methods. The labeled instances can be found in the supplementary material.
We opted for these instances to assess each method with increasing occupancy while avoiding overwhelming the participants when comparing the layouts.}    

\revB{We have designed the user study based on the psychophysical technique of paired comparisons \cite{Tsukida2011}. Specifically, we utilized the two-alternative forced choice (2AFC) paradigm. Each participant was sequentially presented with all possible label layout pairs where, in each pair, the label layouts of the same instance were used and created with different methods. The participants' task was to choose their preferred label layout for each pair in the sequence. To mitigate the learning effect and fatigue, we randomized both the order of the pairs in the sequence and the positions of label layouts (left or right) in pairs.}
\revB{The user study was conducted with 21 participants, consisting of 19 males and two females, with an average age of 21.53 years (ranging from 21 to 24). The average \revDg{experiment} completion time\revDr{for the experiment} was 6 minutes and 44 seconds, with participants taking anywhere from 2 minutes and 54 seconds to 12 minutes and 32 seconds. Two of the 21 participants
were removed using the outlier analysis tool from P\'{e}rez-Ortiz and Mantiuk~\cite{Perez2017}, as their results deviated significantly from the others.}

\revB{We stored the \revDr{data from the experiment}\revDg{choices} in the count matrix $\mathbf{C}$ for each participant individually. Each element $c_{ij}$ in the matrix indicates the number of times that method $i$ was selected over method $j$. We transformed the per-participant-count matrices $\mathbf{C}$ into a quality score (z-score) scale and calculated statistical significance using a customized MATLAB framework \cite{Perez2017}.}
\revB{To transform the \revDr{count} matrix $\mathbf{C}$ to the quality score scale, we used  Thurstone's Law of Comparative Judgment model concerning Case V \cite{Perez2017, Tsukida2011}. 
We employed a Two-tailed test with a significance level of $\alpha = 0.05$ to reject the null hypothesis, ``there is no clear user preference among the tested methods.''}

\begin{figure}[!t]
    \includegraphics[width=\columnwidth]{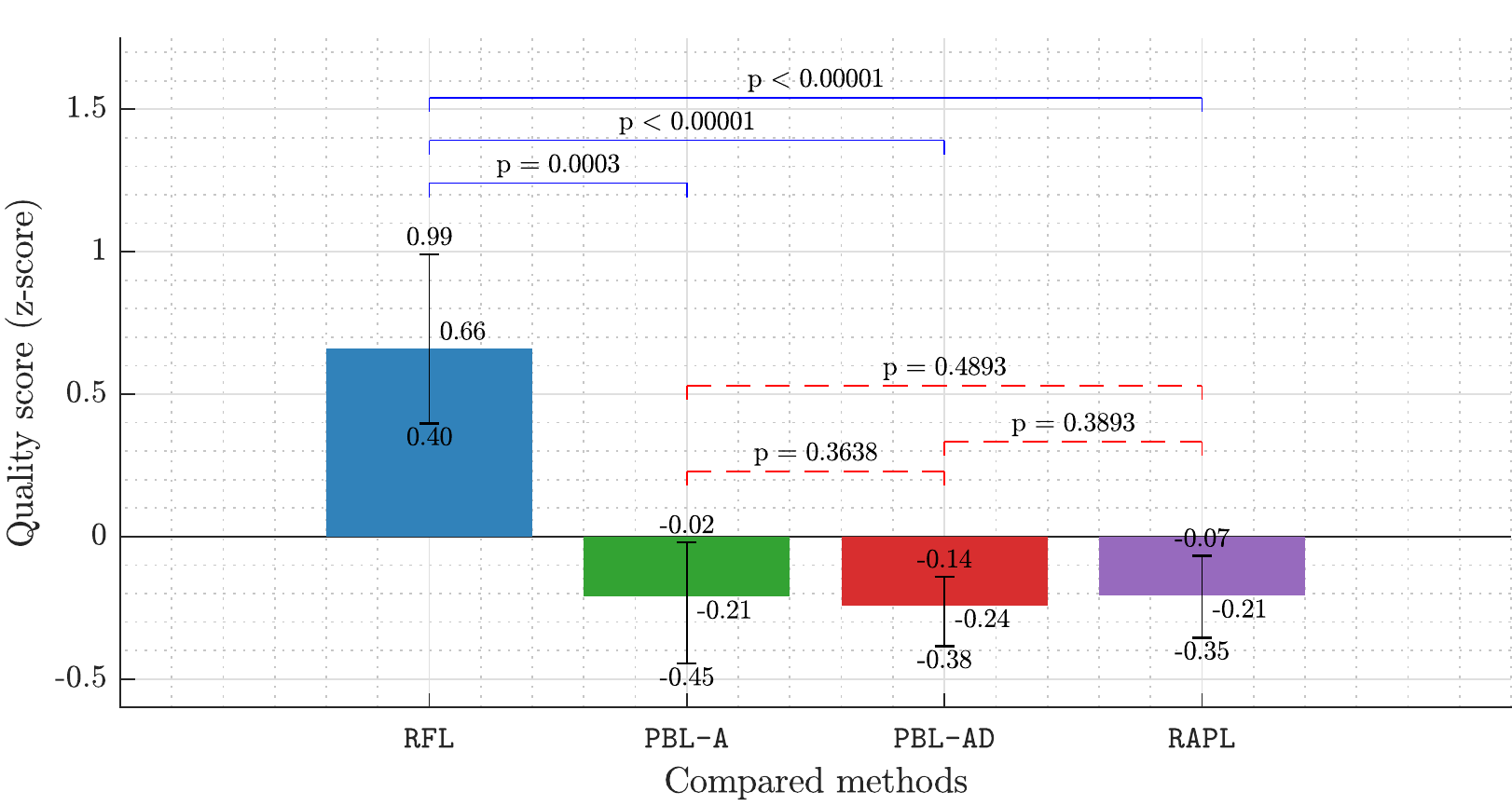}
    \caption{Results of the user study. \revC{The chart shows quality scores accompanied by 95\% confidence intervals. Statistically significant differences between \revDp{the} pairs of methods are denoted by solid blue brackets, with \revDp{the} corresponding $p$-values reported above them. Conversely, \revDp{the} red dashed brackets represent \revDp{the} pairs of methods without evidence of statistically significant differences.}}
    \label{fig:afc-scale-significance}
\end{figure}

\revB{
\revC{\autoref{fig:afc-scale-significance} presents the quality scores and the statistical significance of the
evaluated methods.} 
The results show that the null hypothesis is clearly rejected as the proposed \texttt{RFL} method exhibits the best quality score that is significantly better than the quality scores of the remaining evaluated methods. In other words, the proposed \texttt{RFL} method was preferred by the users over the remaining evaluated methods. The results also suggest that there is not a significant difference in user preference for the remaining \texttt{RAPL}, \texttt{PBL-A}, and \texttt{PBL-AD} methods.}

\subsection{Discussion}
\label{subsec:discussion}
The outcome of the \revDr{evaluation}\revDg{comparison} is manifold. First, we show that our RL-based method achieves an impressive level of generalization. We \revDp{remind} that we trained \texttt{RFL} on random instances of just two anchors and evaluated the method on unseen instances with up to \revA{600} anchors.
\revC{%
The scalability of our method to handle hundreds of agents relies on two key aspects: the design of our environment and the use of Multi-Agent Deep Reinforcement Learning (MADRL). As the number of agents increases, more potential conflicts can occur. However, our local-global reward structure motivates the agents to minimize these conflicts collectively. Notably, our design does not involve any explicit communication channel among the agents. Instead, \revDp{a} shared policy implicitly encourages collaborative behavior, resulting in predictably coordinated actions among agents.}
Even so, \texttt{RFL} outperformed the compared methods in the category of completeness\revB{, and at the same time, the user study participants preferred \texttt{RFL} over all compared methods.} By this fact, we demonstrate the power of machine learning techniques and their capability to surpass the hand-crafted algorithms\revDr{ designed by human experts}.

Second, we show that greedy methods frequently produce suboptimal solutions concerning completeness.  
The quantitative results shown in \autoref{subsec:quantitative-results} and the comparison of \revDp{the} examined methods in \autoref{tab:visual-comparison} provide evidence supporting this claim. 
For example, \revBa{in} instance 45, \texttt{RAPL} and \texttt{PBL-A} left four anchors unlabeled. Even \texttt{PBL-AD}, with the benefit of distant labels, did not find complete labeling and left two anchors unlabeled. 
In contrast, \texttt{RFL} produced complete adjacent labelings for all these instances.

Third, we observe a trade-off between optimality and computation time demands. All the previous methods we examined, \texttt{PBL-A}, \texttt{PBL-AD}, and \texttt{RAPL}, can be computed faster but at the expense of an incomplete solution. In contrast, the proposed \texttt{RFL} method is several orders of magnitude slower but, on the other hand, provides results with \revDg{a} much higher level of completeness. Therefore, the examined methods, \texttt{PBL-A}, \texttt{PBL-AD}, and \texttt{RAPL}, are suitable for interactive applications where incompleteness is not critical (interactive visualizations with the ability to zoom). On the other hand, \texttt{RFL} is better suited for cases where the labeling can be computed in advance, and completeness is essential (\eg cartographic maps, technical drawings, medical atlases). We argue that our approach can serve better than \revDp{the} other examined methods to aid professional illustrators. \revC{We believe that future research based on \texttt{RFL} can further mitigate the gap between optimality and speed.}

\revC{%
Finally, even though we have primarily focused on the elimination of label overlap in this work, the RL framework is much more versatile. It enables the integration of other metrics into the reward function, allowing one to tailor the solution to specific tasks and opening up numerous opportunities for further\revDr{potential} improvements and research. The flexibility solidifies the potential of RL in solving complex spatial decision-making tasks like label placement, promising exciting advancements in the visualization field.}

\section{Limitations}

We are aware of several flaws and limitations of the proposed method. First and foremost, \texttt{RFL} is currently limited to finding only binary solutions. As a result, the method can find either complete conflict-free labeling or a complete but conflict-present solution (\ie label-label or label-anchor at the screen bound). However, the flexible \revDr{definition}\revDg{design} of the \texttt{AdjacentPFLEnv} environment allows one to define new actions or entirely redefine \revDp{the} existing ones. In future work, adding a further indicator action, \revDr{indicating}\revDg{which decide} whether to place the label or not, could \revDr{solve}\revDg{address} the binary limitation of the current method. Similarly, the reward objectives could be extended to minimize the number of unlabeled anchors. \revDr{Furthermore, we plan to release the \texttt{AdjacentPFLEnv} publicly to endorse the research further.}\revDg{To this end, we intend to make the \texttt{AdjacentPFLEnv} publicly available to further support the research in this domain.}

Computation time is another area for improvement of the proposed method. We showed in \autoref{subsec:quantitative-results} that \texttt{RFL} is magnitudes of order slower than \revDp{the} compared methods. Further examination revealed that the computation and collection of observations contribute significantly to the computation time, mainly due to expensive ray casting and imperfect code optimization. However, we believe that in future work, the limitation can be addressed, for instance, by using a graph representation of possible conflicts known as a conflict graph in the observation. Nevertheless, we leave the question open for future research.

\section{Conclusion}
In this work, we introduced the first Multi-Agent Deep Reinforcement Learning formulation of the adjacent-point-feature labeling problem. To facilitate the label placement policy training, we developed \texttt{AdjacentPFLEnv}, an environment where agents collect experiences -- sense the state of the environment via proposed observation modalities, perform actions, and receive feedback in the form of \revBa{proposed} reward. Furthermore, we designed an efficient yet straightforward feedforward neural network architecture with  less than half of a million parameters to model the agent's policy and estimate the value function.
We show that our approach significantly outperforms previous hand-crafted methods designed by human experts in the number of placed labels \revB{and perceived quality}.
Additionally, we would like to encourage the labeling community towards standardized evaluation, a long-used machine learning practice. To this end, we are proposing a new  benchmark dataset to facilitate \revDg{the} comparison of label placement methods, as most of the method codes remain unpublished or proprietary.

\ifCLASSOPTIONcompsoc
  \section*{Acknowledgments}
\else
  \section*{Acknowledgment}
\fi

This work was supported by project \emph{LTAIZ19004 Deep-Learning Approach to Topographical Image Analysis}; 
by the Ministry of Education, Youth and Sports of the Czech Republic within the activity INTER-EXCELENCE (LT), subactivity
INTER-ACTION (LTA), ID: SMSM2019LTAIZ and by Grant Agency of CTU in Prague grant No. SGS22/173/OHK3/3T/13 - Research of Modern Computer Graphics Methods 2022-2024.
Computational resources were mainly supplied by the project \emph{"e-Infrastruktura CZ" (e-INFRA CZ ID:90140)} supported by the Ministry of Education, Youth and Sports of the Czech Republic.

\ifCLASSOPTIONcaptionsoff
  \newpage
\fi

\bibliographystyle{IEEEtran}

\bibliography{references}

\begin{IEEEbiography}
[{\includegraphics[width=1in,height=1.25in,clip,keepaspectratio]{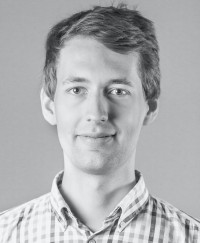}}]{Petr Bob\'{a}k}
is a PhD candidate at the Faculty of Information Technology at Brno University of Technology, Czechia. He received his Master's degree in Computer Science from the same institution in 2017. His research interests include information visualization, machine learning, computer vision, and mathematical optimization.
\end{IEEEbiography}

\begin{IEEEbiography}[{\includegraphics[width=1in,height=1.25in,clip,keepaspectratio]{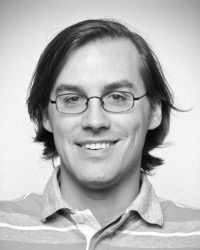}}]{Ladislav \v{C}mol\'{\i}k}
is an assistant professor at the Department of Computer Graphics and Interaction of the Czech Technical University in Prague, Czechia. He received his PhD from the same institution in 2011. His research interests include illustrative visualization, non-photorealistic rendering, and HCI.
\end{IEEEbiography}

\begin{IEEEbiography}[{\includegraphics[width=1in,height=1.25in,clip,keepaspectratio]{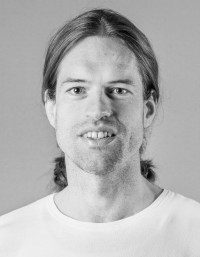}}]{Martin \v{C}ad\'{\i}k} is an associate professor of computer science at Brno University of Technology, 
where he heads his Computational Photography group. He received his PhD from \revDp{the} Czech Technical University in Prague in 2002. His research includes high dynamic range imaging, image processing, computer vision, human visual perception and image, and video quality assessment, among others.\end{IEEEbiography}

\vfill

\end{document}